\pdfoutput=1
\documentclass[10pt,twocolumn,letterpaper]{article}
\usepackage[pagenumbers]{cvpr}

\usepackage{graphicx}
\usepackage{amsmath}
\usepackage{amssymb}
\usepackage{pifont}
\newcommand{\cmark}{\ding{51}}
\newcommand{\xmark}{\ding{55}}
\usepackage{bm}
\usepackage{booktabs}
\usepackage{multirow}
\usepackage{array}
\usepackage{xcolor}
\usepackage{colortbl}
\usepackage{enumitem}
\usepackage{xspace}
\usepackage{placeins}
\usepackage{dblfloatfix}

\makeatletter\def\@flsetnum#1{}\makeatother
\definecolor{best}{RGB}{198,222,227}
\definecolor{second}{RGB}{230,240,242}
\definecolor{cvprblue}{rgb}{0.21,0.49,0.74}
\usepackage[pagebackref,breaklinks,colorlinks,allcolors=cvprblue]{hyperref}
\graphicspath{{figures/}}
\AtBeginEnvironment{table}{}
\AtBeginEnvironment{table*}{}

\def\confName{CVPR}
\def\confYear{2027}

\newcommand{\method}{VG3R\xspace}
\newcommand{\vparm}{Wan-LoRA\xspace}

\providecommand{\ablbShiftMax}{}\renewcommand{\ablbShiftMax}{1.1}
\providecommand{\ablbCamE}{}\renewcommand{\ablbCamE}{8.8}
\providecommand{\ablbCamA}{}\renewcommand{\ablbCamA}{13.1}
\providecommand{\ablbShiftE}{}\renewcommand{\ablbShiftE}{85.6}
\providecommand{\ablbShiftA}{}\renewcommand{\ablbShiftA}{92.4}
\providecommand{\ablbNoShiftE}{}\renewcommand{\ablbNoShiftE}{19.6}
\providecommand{\ablbNoShiftA}{}\renewcommand{\ablbNoShiftA}{25.4}
\providecommand{\ablbConstE}{}\renewcommand{\ablbConstE}{53.2}
\providecommand{\ablbConstA}{}\renewcommand{\ablbConstA}{75.2}
\providecommand{\shiftMin}{}\renewcommand{\shiftMin}{0.08}
\providecommand{\shiftMax}{}\renewcommand{\shiftMax}{1.43}
\newcommand{\cmNoTapsFullAF}{60.4}
\newcommand{\cmNoTapsFullARot}{7.78}
\newcommand{\cmNoTapsFullEF}{61.6}
\newcommand{\cmNoTapsFullERot}{7.80}
\newcommand{\cmTapsConstAF}{55.2}
\newcommand{\cmTapsConstEF}{44.3}
\newcommand{\cmTapsFullAF}{64.4}
\newcommand{\cmTapsFullARot}{3.41}
\newcommand{\cmTapsFullEF}{65.0}
\newcommand{\cmTapsFullERot}{3.79}
\newcommand{\cmTapsNoShiftAF}{21.8}
\newcommand{\cmTapsNoShiftEF}{19.6}

\title{VG3R: Video Generation Model for Feed-Forward 3D Reconstruction}

\author{Jingong Chen$^{1,3\dagger*}$\quad Qingwen Zhang$^{2\dagger*}$\quad Linlian Jiang$^{4,5}$\quad Xinxin Zuo$^{4}$\quad Yang Wang$^{4,5}$\\
Chulwoo Pack$^{1}$\quad Kyle Gao$^{3}$\quad Patric Jensfelt$^{2\ddagger}$\quad Kwanghee Won$^{1\ddagger*}$\\[4pt]
{\normalsize $^{1}$South Dakota State University\quad $^{2}$KTH Royal Institute of Technology\quad $^{3}$Aalto University}\\
{\normalsize $^{4}$Concordia University\quad $^{5}$Mila -- Quebec AI Institute}\\[3pt]
{\normalsize $^{\dagger}$Equal contribution\quad $^{\ddagger}$Equal advising\quad $^{*}$Corresponding author}}

\begin{document}
\makeatletter
\twocolumn[%
  \@maketitle
  \begin{center}
  \captionsetup{hypcap=false}%
  {\includegraphics[width=\textwidth]{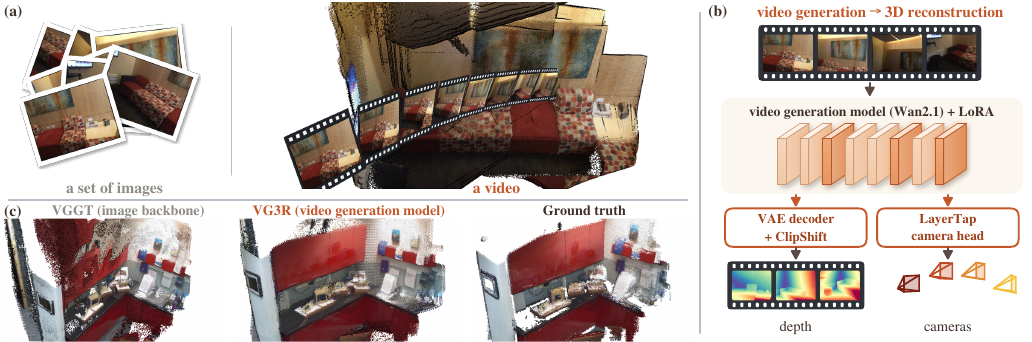}}
  \vspace{-1.5em}
  \captionof{figure}{\textbf{\method} turns a pretrained video generation model into a feed-forward model that reconstructs a scene and the camera of every frame from a video. (a) Geometry foundation models treat a video as a set of images, \method as a video. (b) Overview of \method. (c) On a 7-Scenes video, VGGT, which uses an image backbone, distorts the red partition, which \method reconstructs as in the ground truth.}
  \label{fig::teaser}
  \end{center}
]
\makeatother
\begin{abstract}
Video generation models are widely used as 3D priors for novel view synthesis and 3D scene generation.
Fine-tuned for perception, generation models also excel at dense prediction such as depth estimation.
However, methods that apply video generation models to 3D reconstruction either predict depth without cameras or generate geometry and estimate cameras through many iterative denoising steps, which makes them slower.
In this paper, we present \textbf{\method}, which turns a pretrained video generation model into a feed-forward 3D reconstruction model.
Our key observation is that such a model already captures the depth and camera motion of a video, but not in the form required for reconstruction.
We introduce two components: \textit{clip-level shift prediction}, which corrects depth distortion caused by the unknown shift in decoded inverse depth, and a \textit{multi-layer camera head}, which predicts per-frame cameras parameters by combining features across transformer blocks.
Using Wan2.1 as its video backbone, \method reconstructs a scene and estimate the camera of every frame without iterative denoising.
It achieves the lowest ATE and reconstruction errors among methods built on video generation models, runs about 4 and 14 times faster than Aether and Geo4D. \method also demonstrates competitive depth estimation and 3D reconstruction performance compared with geometry foundation models.
\\Project page:\\
\url{https://github.com/KTH-RPL/VG3R}
\end{abstract}
\vspace{-10pt}

\vspace{-0.5em}
\section{Introduction}
\label{sec::intro}

\begin{figure*}[t]
\centering
\includegraphics[width=\textwidth]{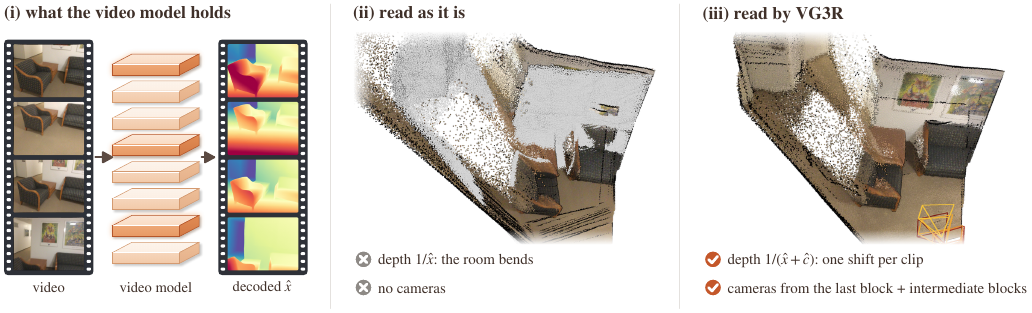}
\vspace{-2em}
\caption{\textbf{Motivation.}
(i) A video generation model used for feed-forward depth prediction outputs clip-normalized inverse depth $\hat{x}$ (shown on a ScanNet clip).
(ii) Normalization sets the farthest surfaces to zero inverse depth, so directly taking $1/\hat{x}$ places them at infinity and distorts the room, even with ground-truth cameras and scale alignment (gray: ground truth).
(iii) \method predicts a clip-level shift $\hat{c}$ to restore the inverse-depth offset and reconstructs the room using its own predicted cameras. Its camera head also reads the highlighted intermediate transformer blocks in (i).}
\label{fig::motiv}
\vspace{-1.5em}
\end{figure*}

Diffusion models generate realistic images~\cite{rombach2022ldm,dit2023} and videos~\cite{svd2023,cogvideox2025,wan2025},
learning rich visual representations in the process.
Beyond generation, image diffusion models have been adapted for depth and normal estimation, showing strong generalization to unseen scenes~\cite{marigold2024,geowizard2024}.
Recent methods also achieve this through deterministic feed-forward prediction~\cite{e2eft2025,lotus2025,genpercept2025}.
Gabeur \etal~\cite{gabeur2026image} turn an image generator into a generalist perception model by casting vision tasks as RGB image generation, matching or outperforming specialist models in segmentation, metric depth, and normal estimation.
They suggest that image generation pretraining could play a role similar to language-model pretraining in enabling general-purpose perception.
Video generation models go further: they generate videos that remain coherent as camera and objects move, which suggests that they implicitly learn the 3D structure of scenes.
They have likewise been adapted to video depth estimation~\cite{depthcrafter2025,chronodepth2025,depthanyvideo2025,dvd2026}, and GenCeption~\cite{genception2026} turns Wan2.1~\cite{wan2025} into a single-step model that matches or exceeds Depth Anything 3~\cite{depthanything3_2026}, SAM~3~\cite{sam3_2025} and VGGT-$\Omega$~\cite{vggtomega2026} on their own tasks with less training data.
Many works also use the 3D prior of video generation models, from objects to scenes: they generate multi-view videos of objects~\cite{sv3d2024,sv4d2025}, synthesize novel views of scenes~\cite{viewcrafter2025,reconx2026} and create world-consistent videos and explorable 3D scenes~\cite{gen3c2025,voyager2025,lyra2026}.

However, directly recovering both scene geometry and camera poses from a video remains challenging for video generation models.
The 3D generation methods above synthesize views along given camera trajectories, whereas 3D reconstruction must recover both geometry and cameras from the input video.
Video depth estimators~\cite{depthcrafter2025,chronodepth2025,depthanyvideo2025,dvd2026} predict depth or inverse depth up to scale and shift, but require separate camera pose estimation and alignment for 3D reconstruction.
GenCeption~\cite{genception2026} predicts one output per prompt, such as depth or a ray map, not depth and cameras jointly.
Methods that jointly generate geometry and estimate cameras with video generation models~\cite{aether2025,geo4d2025,sora3r2025,gen3r2026} need many iterative denoising steps, which makes them slower.
Consequently, feed-forward reconstruction is dominated by geometry foundation models~\cite{dust3r2024,mast3r2024,monst3r2025,cut3r2025} such as VGGT~\cite{vggt2025}, $\pi^3$~\cite{pi3_2026}, MapAnything~\cite{mapanything2026}, VGGT-$\Omega$~\cite{vggtomega2026} and Depth Anything 3~\cite{depthanything3_2026}.
These models encode each frame with an image backbone such as DINOv2~\cite{dinov2_2024} and treat a video as a set of images (\cref{fig::teaser}(a), left).

In this paper, we show that a video generation model can be turned directly into a feed-forward reconstruction model.
Our key observation is that, used as a feed-forward network, a video generation model already captures the depth and camera motion of a video, but not in the form required for reconstruction (\cref{fig::motiv}).
For depth, the deterministic depth pipeline of DVD~\cite{dvd2026} decodes the inverse depth $\hat{x}$ of every frame from the video diffusion transformer without sampling (\cref{fig::motiv}(i)).
However, $\hat{x}$ is normalized over the clip to $[0,1]$, from $0$ at the farthest surfaces to $1$ at the nearest, so taking $1/\hat{x}$ as depth places the farthest surfaces at infinity and exaggerates the depth of far surfaces relative to near ones.
Even with the ground-truth cameras and scale, the room reconstructed this way bends away from the ground truth (\cref{fig::motiv}(ii)).
Recovering depth requires estimating, for each clip, the shift $c$ removed by this normalization, such that the inverse depth is proportional to $\hat{x}+c$~\cite{leres2021,moge2025,moge2_2025}.
This shift is the inverse depth of the farthest surfaces divided by the range of inverse depth in the clip, so it depends on the relative depths of the nearest and farthest surfaces, while the remaining unknown scale only changes the overall size of the reconstruction.
Beyond depth, reconstructing the scene also requires camera poses. The transformer features contain information about camera motion, but existing depth pipeline does not estimate cameras (\cref{fig::motiv}(ii)).

Based on this observation, we present \method, which fine-tunes Wan2.1~\cite{wan2025} with LoRA~\cite{lora2022} into a feed-forward model that reconstructs a scene and estimates the cameras of all frames from a video (\cref{fig::teaser}(a), right), without iterative denoising or a separate pose estimator (\cref{fig::teaser}(b) and \cref{fig::motiv}(iii)).
Building on DVD's deterministic depth pipeline, \method introduces two components to recover depth and cameras jointly.
For depth, clip-level shift prediction (ClipShift) pools transformer features across the video to predict a single shift, correcting the decoded inverse depth while preserving one global scale ambiguity.
For cameras, the multi-layer camera head (LayerTap) combines features from intermediate and final transformer blocks to predict the camera parameters of every frame.
The point map of every frame then follows from its depth and camera.
On standard video benchmarks, \method achieves the lowest ATE on every dataset and the lowest reconstruction errors among the methods built on video generation models, and runs about 4 and 14 times faster than Aether~\cite{aether2025} and Geo4D~\cite{geo4d2025}.
Compared with VGGT, it reduces video depth error by $14\%$ to $33\%$ and reconstruction errors on 7-Scenes (\cref{fig::teaser}(c)), demonstrating competitive performance with geometry foundation models.

\noindent
In summary, our contributions are as follows:
\begin{itemize}[leftmargin=*]
\itemsep2pt
\item We present \method, a feed-forward 3D reconstruction model fine-tuned from a video generation model, which predicts a scene and the camera of every frame from a video without iterative denoising and thus establishes a direct path from video generation to 3D reconstruction.
\item We find that a video generation model already captures the depth and camera motion of a video, and propose ClipShift and LayerTap to make them available for reconstruction.
\item We show that \method outperforms the methods built on video generation models in ATE and reconstruction errors, approaches geometry foundation models and runs about 4 and 14 times faster than Aether and Geo4D.
\end{itemize}

\section{Related Work}
\label{sec::related}

\begin{figure*}[t]
\centering
\includegraphics[width=\textwidth]{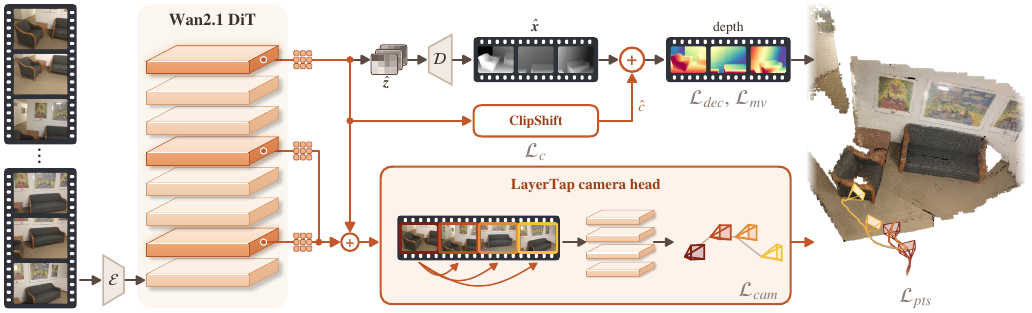}
\vspace{-1.5em}
\caption{\textbf{Overview.}
\method encodes the input video with the frozen VAE of Wan2.1 and processes its latents with a LoRA-adapted transformer at a fixed timestep, with blocks ordered from bottom to top.
The final-block features are decoded into inverse depth $\hat{\bm{x}}$ through the output layer and VAE decoder $\mathcal{D}$.
ClipShift uses these features to predict a single shift $\hat{c}$ for a video clip, recovering depth up to a global scale.
LayerTap combines features from multiple transformer blocks and pairs each latent frame with the first to predict per-frame cameras.
Depth and cameras yield the reconstructed point maps. Gray losses are used only during training.}
\label{fig::method}
\vspace{-1em}
\end{figure*}

\vspace{-2pt}
\subsection{Generative Priors for 3D and Perception}
\vspace{-2pt}
Video generation models are widely used as 3D priors.
They synthesize novel views from one or a few images~\cite{viewcrafter2025,reconx2026}, generate multi-view videos of static and dynamic objects~\cite{sv3d2024,sv4d2025}, enable camera-controlled generation~\cite{cameractrl2025,gen3c2025} and create explorable 3D scenes~\cite{voyager2025,lyra2026,lyra2_2026}.
Image diffusion models have also been adapted for depth and normal estimation~\cite{marigold2024,marigoldjournal2025,geowizard2024}, including deterministic single-step variants~\cite{lotus2025,genpercept2025,e2eft2025}.
Vision Banana~\cite{gabeur2026image} instruction-tunes an image generator into a generalist vision model by casting the outputs of vision tasks as RGB images.
Unlike these works, \method uses a video generation model as a feed-forward 3D reconstruction model rather than for generation or depth estimation alone.

\vspace{-2pt}
\subsection{Video Generation Models for Geometry}
\vspace{-2pt}
Video generation models have been adapted to predict geometry in several ways, ranging from direct regression to iterative generation.
GenCeption~\cite{genception2026} adapts Wan2.1~\cite{wan2025} for single-step perception, using text prompts to select tasks such as depth or ray map estimation.
Built on Stable Video Diffusion~\cite{svd2023}, DepthCrafter~\cite{depthcrafter2025}, ChronoDepth~\cite{chronodepth2025} and Depth Any Video~\cite{depthanyvideo2025} predict affine-invariant depth or disparity, while GeometryCrafter~\cite{geometrycrafter2025} and MotionCrafter~\cite{motioncrafter2026} predict point maps.
DVD~\cite{dvd2026}, whose depth pipeline we adopt, adapts Wan2.1-T2V-1.3B into a deterministic regressor from the RGB latent at a fixed timestep.
Geo4D~\cite{geo4d2025}, Aether~\cite{aether2025}, Sora3R~\cite{sora3r2025} and Gen3R~\cite{gen3r2026} generate geometry and cameras through iterative  denoising.
Other works obtain geometry through next-frame prediction~\cite{geonext2026}, the prediction heads of a reconstruction model~\cite{latentbridge2026}, or joint modeling with other modalities~\cite{omnivdiff2026,unityvideo2026,diffusionrenderer2025}.

Beyond explicit geometry prediction, probing studies examine the 3D information encoded in frozen features of image, video and vision-language models~\cite{probe3d2024,lexicon3d2024,probevlmvgm2026}.
VidFM3D~\cite{vidfm3d2026} finds that frozen Wan2.1 features outperform DINO features for reconstruction, while Geometry Forcing~\cite{geometryforcing2026} aligns video diffusion features with a 3D foundation model.
Unlike methods that predict only depth or rely on iterative generation, \method
jointly predicts depth and cameras with a deterministic feed-forward network.
Concurrent with our work, Gen4P~\cite{gen4p2026} also fine-tunes Wan2.1 with a rank-512 LoRA into a feed-forward model, predicting video depth and camera poses alongside 3D point tracking.
Gen4P uses a separate cross-attention decoder to answer queries, recovers cameras from queried points through Kabsch fits and pose-graph optimization, and can incorporate DVD's LoRA as a frozen prior.
In contrast, \method decodes depth with the video generation model's VAE and directly regresses cameras from its transformer features.

Without a video generation prior, CVD~\cite{cvd2020} and MegaSaM~\cite{megasam2025} optimize consistent depth for each video at test time, RollingDepth~\cite{rollingdepth2025} aligns image diffusion predictions across short video snippets, and Video Depth Anything~\cite{videodepthanything2025} adds a temporal head to Depth Anything V2~\cite{depthanythingv2_2024}.
Inspired by CVD's geometric consistency constraints, \method instead enforces cross-frame depth consistency during training through clip-level depth alignment and multi-view supervision using ground-truth correspondences.

\vspace{-2pt}
\subsection{3D Geometry Foundation Models}
\vspace{-2pt}
Geometry foundation models learn feed-forward 3D reconstruction from multiple views.
DUSt3R~\cite{dust3r2024} trained one network to regress the point maps of an image pair in a common coordinate frame, and MASt3R~\cite{mast3r2024} added a matching head to the same backbone.
MonST3R~\cite{monst3r2025} and CUT3R~\cite{cut3r2025} extend this to dynamic scenes and to streams of frames.
VGGT~\cite{vggt2025} predicts the cameras, depth and point maps of many views with one transformer, $\pi^3$~\cite{pi3_2026} removes its reference view so that the prediction is equivariant to the order of the views, and MapAnything~\cite{mapanything2026} accepts optional inputs such as intrinsics and predicts metric scale.
All of them build on image backbones~\cite{dinov2_2024} and train on many multi-view datasets with 3D supervision.
\method retains their feed-forward design and replaces the image backbone with a video generation model pretrained on frames in temporal order.

\vspace{-3pt}
\section{Method}
\vspace{-3pt}
\label{sec::method}
Given a video clip $\bm{I}=\{\bm{I}_i\}_{i=0}^{L-1}$ of $L$ frames, \method predicts the depth and the camera of every frame, recovering the scene as point maps up to a global scale.
Building on the deterministic depth pipeline of DVD~\cite{dvd2026} (\cref{sec::prelim}), \method adds ClipShift to recover depth up to scale (\cref{sec::shift}) and LayerTap to predict per-frame camera parameters (\cref{sec::cam}).
\Cref{sec::train} describes the training objectives and processing of long videos.

\subsection{Preliminaries}
\label{sec::prelim}

The frozen Wan VAE~\cite{vae2013,wan2025}, with encoder $\mathcal{E}$ and decoder $\mathcal{D}$, encodes the first frame alone and every following group of four frames into one latent frame at $1/8$ of the spatial resolution, so that the latent $\bm{z}_I=\mathcal{E}(\bm{I})$ has $N=1+(L-1)/4$ latent frames.
Wan2.1-T2V-1.3B~\cite{wan2025} generates videos with a diffusion transformer (DiT)~\cite{dit2023} of $30$ blocks trained by flow matching~\cite{flowmatching2023,rectifiedflow2023}.
DVD turns this model into a deterministic depth regressor that maps the noise-free video latent to a depth latent $\hat{\bm z}$ at a fixed timestep $\tau_0$,
\begin{equation}
\label{eq::forward}
\hat{\bm{z}}=F_\theta\bigl(\bm{z}_I;\tau_0,\varnothing\bigr),\qquad
\hat{\bm{x}}=\tfrac12\bigl(\bar{\mathcal{D}}(\hat{\bm{z}})+1\bigr),
\end{equation}
where $F_\theta$ is the DiT with frozen pretrained weights and trainable LoRA~\cite{lora2022} weights $\theta$, $\varnothing$ is the empty text prompt, and $\bar{\mathcal{D}}$ averages the three output channels of the VAE decoder $\mathcal{D}$.
The target of $\hat{\bm{x}}$ is the inverse depth $\delta=1/d$ under the percentile normalization of Marigold~\cite{marigold2024}, which DVD applies over the whole input video clip,
\begin{equation}
\label{eq::target}
x_i(\bm{p})=\frac{\delta_i(\bm{p})-\delta_{\mathrm{lo}}}{\delta_{\mathrm{hi}}-\delta_{\mathrm{lo}}},
\end{equation}
clipped to $[0,1]$, where $\bm{p}$ is a pixel and $\delta_{\mathrm{lo}}$ and $\delta_{\mathrm{hi}}$ are the 2nd and 98th percentiles of $\delta$ over the valid pixels of the video clip.
DVD supervises $\hat{\bm{z}}$ with the squared error $\mathcal{L}_{\mathrm{lat}}$ to the encoded target and the error $\mathcal{L}_{\mathrm{grad}}$ of its differences along latent frames, rows and columns.
The DVD pipeline thus has two limitations for 3D reconstruction (\cref{fig::motiv}(ii)): the decoded inverse depth $\hat{\bm{x}}$ requires a shift to recover the scene geometry up to scale, and the model does not estimate camera intrinsics and extrinsics.

\subsection{ClipShift: Clip-Level Shift Prediction}
\label{sec::shift}

The decoded inverse depth $\hat{\bm{x}}$ cannot be used directly for 3D reconstruction, as the missing shift distorts the scene geometry (\cref{fig::motiv}(ii)).
Recovering depth requires adding a clip-level shift $c$ to $\hat{\bm{x}}$, giving depth proportional to $1/(\hat{\bm{x}}+c)$.
From \cref{eq::target}, the corresponding ground-truth shift is $c^\ast=\delta_{\mathrm{lo}}/(\delta_{\mathrm{hi}}-\delta_{\mathrm{lo}})$, giving $\delta_i=(\delta_{\mathrm{hi}}-\delta_{\mathrm{lo}})(x_i+c^\ast)$.
The remaining scale factor $\delta_{\mathrm{hi}}-\delta_{\mathrm{lo}}$ only changes the overall size of the reconstruction, while the shift determines its shape.

Importantly, the shift is invariant to scene scale and shared across all frames of the input video clip. This allows us to recover the scene geometry up to scale by predicting a single shift from the video clip. We therefore introduce \textit{clip-level shift prediction} (ClipShift), which pools the tokens of the last DiT block across all latent frames using attention and predicts a positive shift $\hat{c}$ with an MLP. The depth of every frame and the shift loss are
\begin{equation}
\label{eq::depth}
\hat{D}_i(\bm{p})=\frac{1}{\max\bigl(\hat{x}_i(\bm{p})+\hat{c},\,\tau\bigr)},
\quad
\mathcal{L}_{\mathrm{c}}=\Bigl\lvert\ln\frac{\hat{c}+\epsilon}{c^\ast+\epsilon}\Bigr\rvert,
\end{equation}
where $\tau$ and $\epsilon$ are small constants to prevents division by zero and stabilizes the logarithmic loss, respectively.

\begin{table*}[t]\centering\scriptsize
\setlength{\tabcolsep}{1pt}
\caption{\textbf{Comparison on video benchmarks.}
Video depth is aligned once per video, with scale and shift fitted in inverse depth ($^\dagger$) or depth ($^\ddagger$), and scale-only alignment otherwise.
A dash indicates methods without camera predictions.
Bold and underline mark the best and second-best results among \method and video-generation-based methods; dark and light shading mark those among \method and geometry foundation models.
\method achieves the lowest ATE among video-generation-based methods across all three trajectory benchmarks, and outperforms all geometry foundation models in depth error on Bonn and reconstruction completeness error on 7-Scenes.
}
\vspace{-0.8em}
\label{tab::main}
\resizebox{\textwidth}{!}{%
\begin{tabular}{@{}l*{22}{c}@{}}
\toprule
 & \multicolumn{7}{c}{Video depth} & \multicolumn{9}{c}{Camera trajectories} & \multicolumn{6}{c}{Reconstruction}\\
\cmidrule(lr){2-8}\cmidrule(lr){9-17}\cmidrule(lr){18-23}
 & \multicolumn{2}{c}{Sintel} & \multicolumn{2}{c}{Bonn} & \multicolumn{2}{c}{KITTI} & ScanNet & \multicolumn{3}{c}{Sintel} & \multicolumn{3}{c}{TUM-dynamics} & \multicolumn{3}{c}{ScanNet} & \multicolumn{3}{c}{7-Scenes} & \multicolumn{3}{c}{NRGBD}\\
\cmidrule(lr){2-3}\cmidrule(lr){4-5}\cmidrule(lr){6-7}\cmidrule(lr){8-8}\cmidrule(lr){9-11}\cmidrule(lr){12-14}\cmidrule(lr){15-17}\cmidrule(lr){18-20}\cmidrule(lr){21-23}
Method & AbsRel$\downarrow$ & $\delta_{1.25}$$\uparrow$ & AbsRel$\downarrow$ & $\delta_{1.25}$$\uparrow$ & AbsRel$\downarrow$ & $\delta_{1.25}$$\uparrow$ & TAE$\downarrow$ & ATE$\downarrow$ & RPE-t$\downarrow$ & RPE-r$\downarrow$ & ATE$\downarrow$ & RPE-t$\downarrow$ & RPE-r$\downarrow$ & ATE$\downarrow$ & RPE-t$\downarrow$ & RPE-r$\downarrow$ & Acc$\downarrow$ & Comp$\downarrow$ & NC$\uparrow$ & Acc$\downarrow$ & Comp$\downarrow$ & NC$\uparrow$\\
\midrule
\multicolumn{23}{@{}l}{\emph{Built on video generation models}}\\
DepthCrafter$^{\dagger}$~\cite{depthcrafter2025} & 0.305 & 0.615 & 0.072 & 0.973 & 0.104 & 0.896 & 1.606 & -- & -- & -- & -- & -- & -- & -- & -- & -- & -- & -- & -- & -- & -- & --\\
Depth Any Video$^{\dagger}$~\cite{depthanyvideo2025} & 0.304 & 0.641 & 0.072 & 0.968 & 0.072 & 0.958 & 1.998 & -- & -- & -- & -- & -- & -- & -- & -- & -- & -- & -- & -- & -- & -- & --\\
DKT$^{\dagger}$~\cite{dkt2025} & 0.479 & 0.457 & 0.154 & 0.779 & 0.206 & 0.678 & 1.420 & -- & -- & -- & -- & -- & -- & -- & -- & -- & -- & -- & -- & -- & -- & --\\
DVD$^{\dagger}$~\cite{dvd2026} & 0.268 & \underline{0.672} & 0.058 & 0.972 & 0.064 & 0.970 & 1.143 & -- & -- & -- & -- & -- & -- & -- & -- & -- & -- & -- & -- & -- & -- & --\\
ChronoDepth$^{\ddagger}$~\cite{chronodepth2025} & 0.677 & 0.413 & 0.099 & 0.917 & 0.168 & 0.754 & 1.599 & -- & -- & -- & -- & -- & -- & -- & -- & -- & -- & -- & -- & -- & -- & --\\
GeometryCrafter~\cite{geometrycrafter2025} & 0.302 & 0.668 & \underline{0.055} & \underline{0.974} & \textbf{0.055} & \textbf{0.984} & 1.238 & 0.230 & 0.077 & 0.926 & 0.072 & \underline{0.019} & \textbf{0.415} & \underline{0.078} & \underline{0.020} & \textbf{0.530} & \underline{0.047} & \underline{0.025} & \textbf{0.609} & \underline{0.065} & \underline{0.019} & \underline{0.660}\\
GenCeption~\cite{genception2026} & \textbf{0.231} & 0.662 & 0.061 & 0.972 & 0.059 & \underline{0.979} & 1.611 & -- & -- & -- & -- & -- & -- & -- & -- & -- & -- & -- & -- & -- & -- & --\\
Aether~\cite{aether2025} & 0.270 & 0.600 & 0.249 & 0.669 & 0.117 & 0.896 & 1.136 & \underline{0.174} & \underline{0.054} & 0.684 & 0.091 & \textbf{0.012} & 1.121 & 0.177 & 0.028 & 1.222 & 0.100 & 0.077 & 0.557 & 0.127 & 0.040 & 0.604\\
Geo4D~\cite{geo4d2025} & 0.335 & 0.574 & 0.088 & 0.916 & 0.091 & 0.937 & \underline{1.052} & 0.179 & 0.062 & \underline{0.503} & \underline{0.071} & \underline{0.019} & 0.637 & 0.123 & 0.022 & 0.700 & 0.049 & 0.033 & \underline{0.597} & 0.071 & 0.026 & 0.632\\
\midrule
\textbf{\method (Ours)} & \cellcolor{second}\underline{0.254} & \cellcolor{best}\textbf{0.699} & \cellcolor{best}\textbf{0.037} & \cellcolor{best}\textbf{0.979} & \underline{0.058} & \underline{0.979} & \cellcolor{second}\textbf{0.734} & \textbf{0.142} & \cellcolor{second}\textbf{0.034} & \textbf{0.360} & \textbf{0.031} & \textbf{0.012} & \underline{0.457} & \textbf{0.035} & \textbf{0.015} & \underline{0.536} & \textbf{0.018} & \cellcolor{best}\textbf{0.017} & \textbf{0.609} & \textbf{0.032} & \textbf{0.015} & \textbf{0.666}\\
\midrule
\multicolumn{23}{@{}l}{\emph{Geometry foundation models}}\\
VGGT~\cite{vggt2025} & 0.296 & 0.647 & 0.055 & 0.971 & 0.072 & 0.965 & 0.848 & 0.171 & 0.061 & 0.471 & \cellcolor{second}0.012 & \cellcolor{second}0.010 & \cellcolor{best}0.309 & 0.035 & 0.015 & 0.376 & 0.021 & 0.024 & 0.630 & \cellcolor{second}0.010 & \cellcolor{second}0.011 & 0.803\\
$\pi^3$~\cite{pi3_2026} & \cellcolor{best}0.232 & 0.665 & 0.050 & \cellcolor{second}0.975 & \cellcolor{best}0.039 & \cellcolor{second}0.986 & \cellcolor{best}0.626 & \cellcolor{best}0.073 & 0.039 & \cellcolor{second}0.281 & 0.014 & \cellcolor{best}0.009 & \cellcolor{second}0.311 & \cellcolor{best}0.030 & \cellcolor{second}0.013 & \cellcolor{best}0.347 & \cellcolor{best}0.012 & \cellcolor{second}0.019 & \cellcolor{best}0.651 & \cellcolor{second}0.010 & \cellcolor{second}0.011 & \cellcolor{second}0.819\\
MapAnything~\cite{mapanything2026} & 0.337 & 0.580 & 0.063 & 0.972 & 0.074 & 0.972 & 1.095 & 0.230 & 0.082 & 0.644 & 0.028 & 0.019 & 0.385 & 0.060 & 0.022 & 0.617 & 0.030 & 0.029 & 0.603 & 0.048 & 0.027 & 0.666\\
DA3~\cite{depthanything3_2026} & 0.313 & \cellcolor{second}0.668 & \cellcolor{second}0.049 & 0.973 & 0.057 & 0.978 & 1.058 & 0.127 & 0.050 & 0.351 & 0.013 & \cellcolor{best}0.009 & 0.312 & 0.033 & \cellcolor{second}0.013 & 0.386 & \cellcolor{second}0.015 & 0.020 & 0.628 & \cellcolor{best}0.008 & \cellcolor{best}0.010 & \cellcolor{best}0.848\\
VGGT-$\Omega$~\cite{vggtomega2026} & 0.256 & 0.660 & 0.114 & 0.933 & \cellcolor{second}0.040 & \cellcolor{best}0.987 & 0.748 & \cellcolor{second}0.087 & \cellcolor{best}0.033 & \cellcolor{best}0.226 & \cellcolor{best}0.011 & \cellcolor{second}0.010 & 0.316 & \cellcolor{second}0.032 & \cellcolor{best}0.012 & \cellcolor{second}0.352 & 0.021 & 0.028 & \cellcolor{second}0.643 & 0.013 & \cellcolor{second}0.011 & 0.793\\
\bottomrule
\end{tabular}}
\vspace{-1.5em}
\end{table*}

\begin{figure*}[t]
\centering
\includegraphics[width=\textwidth]{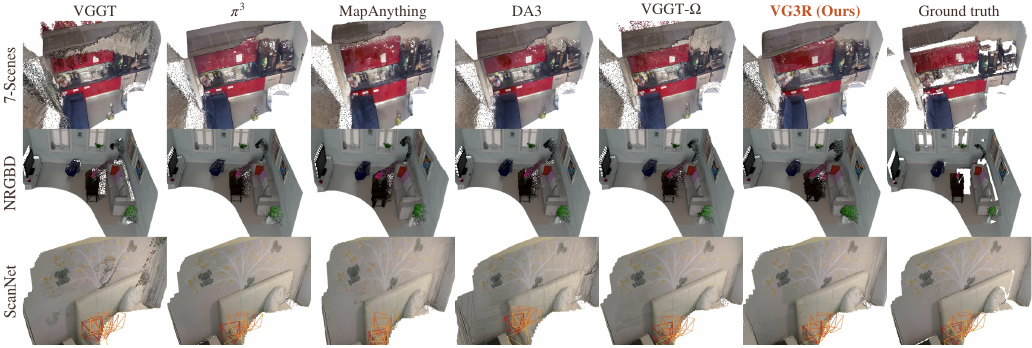}
\vspace{-2em}
\caption{\textbf{Reconstructions from video.} One sequence per row. Every method fuses the point maps of all frames with its own predicted cameras, without confidence filtering or other post-processing, and is aligned to the ground truth by a similarity transform.}
\label{fig::qual}
\vspace{-1em}
\end{figure*}
\subsection{LayerTap: Multi-Layer Camera Head}
\label{sec::cam}

Beyond depth, reconstructing a scene also requires camera intrinsics and extrinsics estimation. Unlike depth, which is decoded from the final transformer block, camera estimation can benefit from features at multiple layers that capture camera motion. LayerTap therefore combines the tokens of intermediate blocks $10$ and $20$ with those of the final block through linear projections and summation (\cref{fig::method}).

To estimate camera poses relative to the first frame, we preserve the spatial tokens of each latent frame and pair them with those of the first latent frame. However, the Wan VAE compresses four RGB frames into each subsequent latent frame, while we need a separate camera estimation for every RGB frame. We use one learned query per RGB frame to extract its features from the corresponding pair of latent frames, yielding $L$ frame-level tokens.
Following VGGT~\cite{vggt2025}, a camera head processes these tokens with self-attention and iteratively refines the rotation $\hat{\bm{R}}_i$, translation $\hat{\bm{t}}_i$, and fields of view $\hat{\bm{f}}_i$ of each frame relative to camera $0$. The ground-truth cameras are expressed in the same reference frame, with translations normalized by the median depth of frame $0$, as for the predicted depth. The camera loss sums over $K$ refinement iterations,
\begin{equation}
\label{eq::cam}
\begin{aligned}
\mathcal{L}_{\mathrm{cam}}=\sum_{k=1}^{K}\gamma^{K-k}\operatorname*{mean}_{i}\Bigl(&w_t\lVert\hat{\bm{t}}^k_i-\bm{t}_i\rVert_1+\lVert\hat{\bm{R}}^k_i-\bm{R}_i\rVert_F\\
&+w_f\lVert\hat{\bm{f}}^k_i-\bm{f}_i\rVert_1\Bigr),
\end{aligned}
\end{equation}
where $\gamma<1$ gives later iterations larger weights, and $w_t$ and $w_f$ balance the three terms.

\subsection{Training Loss and Inference}
\label{sec::train}

\begin{figure*}[t]
\centering
\includegraphics[width=\textwidth]{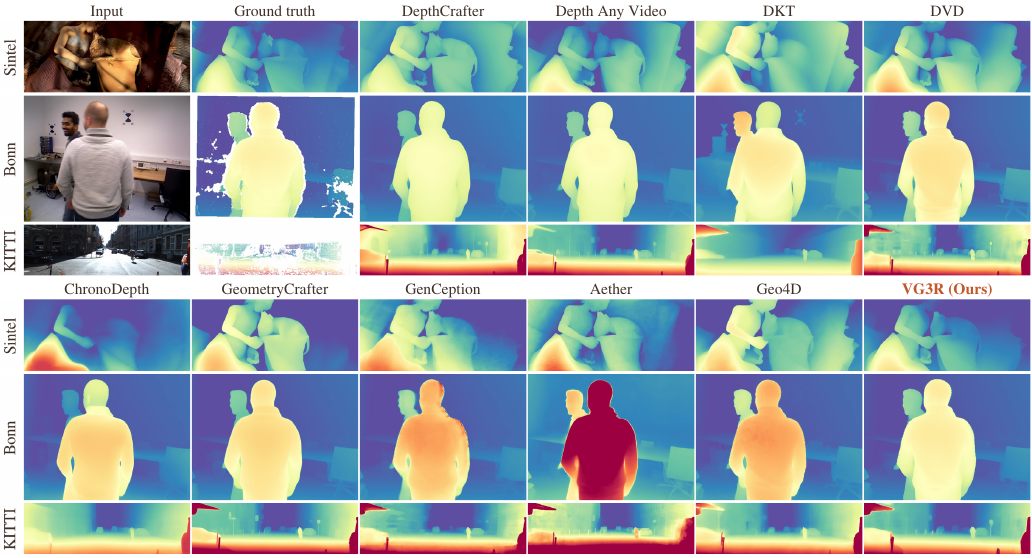}
\vspace{-1.8em}
\caption{\textbf{Video depth from video generation models.} One frame per benchmark, shown as inverse depth after alignment in \cref{tab::main}, using a shared color range with the ground truth for each sequence.}
\label{fig::qualvd}
\vspace{-1.2em}
\end{figure*}

With depth from ClipShift and cameras from LayerTap, \method reconstructs the scene by back-projecting each frame into the camera coordinate system of frame $0$,
\begin{equation}
\label{eq::points}
\hat{\bm{P}}_i(\bm{p})=\hat{\bm{R}}_i^{\top}\bigl(\hat{D}_i(\bm{p})\,\hat{\bm{K}}^{-1}\tilde{\bm{p}}-\hat{\bm{t}}_i\bigr),
\end{equation}
where $\hat{\bm{K}}$ is the camera intrinsic matrix shared across all frames, constructed from the predicted fields of view $\hat{\bm{f}}_0$ of frame $0$ with the principal point at the image center, and $\tilde{\bm{p}}$ denotes the homogeneous coordinates of pixel $\bm{p}$.
We supervise the resulting point maps with an $L_1$ loss $\mathcal{L}_{\mathrm{pts}}$ against ground truth, jointly training the depth, shift, and camera parameter estimation.

To encourage consistent depth across frames, we introduce two additional losses. The clip-level decoded loss $\mathcal{L}_{\mathrm{dec}}$ fits a single affine transformation in inverse depth across all frames and penalizes the remaining log-depth error. Unlike per-frame alignment, this shared fit exposes inconsistencies between frames. The multi-view loss $\mathcal{L}_{\mathrm{mv}}$ enforces depth consistency across frame pairs using correspondences derived from ground-truth depth and cameras. We use pairs at multiple temporal distances to capture both local inconsistencies and longer-range drift. The total training loss is
\begin{equation}
\label{eq::total}
\begin{aligned}
\mathcal{L}={}&\mathcal{L}_{\mathrm{lat}}+0.5\,\mathcal{L}_{\mathrm{grad}}+\mathcal{L}_{\mathrm{dec}}+4\,\mathcal{L}_{\mathrm{mv}}\\
&+\mathcal{L}_{\mathrm{cam}}+\mathcal{L}_{\mathrm{c}}+0.5\,\mathcal{L}_{\mathrm{pts}}.
\end{aligned}
\end{equation}

For inference on videos longer than the training clips, \method processes the whole video and overlapping windows of the training length. The whole-video prediction provides a common depth reference, while the windows provide depth and camera estimates at the training length. Each window's depth is scaled to the whole-video prediction by the median log-depth ratio over their shared frames, and overlapping predictions are blended linearly in log depth. For cameras, each window is aligned to the preceding reconstruction using a rigid transform averaged over the overlapping camera poses.

\section{Experiments}
\label{sec::exp}

\subsection{Setup}
\label{sec::setup}

\begin{figure*}[t]
\centering
\includegraphics[width=\textwidth]{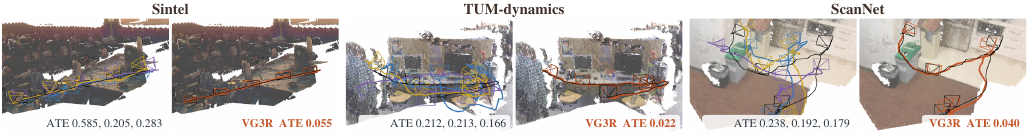}
\vspace{-2.0em}
\caption{\textbf{Camera trajectories} over the ground-truth point cloud, with ground-truth trajectories in black. One sequence per benchmark among the five where \method leads most in ATE. Left: GeometryCrafter (blue), Aether (purple) and Geo4D (yellow), with ATE reported in this order. Right: \method.}
\label{fig::traj}
\vspace{-1.5em}
\end{figure*}
\begin{table}[t]\centering\footnotesize
\setlength{\tabcolsep}{3.5pt}
\caption{\textbf{Inference time} in ms per frame on one GH200, reported as the median across sequences, excluding model loading. Each method uses its official resolution.
}
\vspace{-1em}
\label{tab::time}
\begin{tabular}{@{}lccccc@{}}
\toprule
 & & \multicolumn{3}{c}{Video depth} & Cameras\\
\cmidrule(lr){3-5}\cmidrule(l){6-6}
Method & Steps & Sintel & Bonn & KITTI & ScanNet\\
\midrule
DepthCrafter~\cite{depthcrafter2025} & 5 & 146 & 100 & 158 & --\\
Depth Any Video~\cite{depthanyvideo2025} & 3 & 184 & 133 & 128 & --\\
DKT~\cite{dkt2025} & 5 & 170 & 138 & 117 & --\\
DVD~\cite{dvd2026} & 1 & \textbf{72} & \textbf{64} & 160 & --\\
ChronoDepth~\cite{chronodepth2025} & 5 & 225 & 166 & 168 & --\\
GeometryCrafter~\cite{geometrycrafter2025} & 5 & 310 & 230 & 310 & --\\
GenCeption~\cite{genception2026} & 1 & 216 & 264 & 288 & --\\
Aether~\cite{aether2025} & 4 & 1319 & 1979 & 2958 & 383\\
Geo4D~\cite{geo4d2025} & 5 & 1116 & 1499 & 881 & 1351\\
\midrule
\method & 1 & 74 & 79 & \textbf{78} & \textbf{84}\\
\bottomrule
\end{tabular}
\vspace{-1.5em}
\end{table}

\noindent\textbf{Training.} We train \method on $11$ public datasets: ScanNet++~\cite{scannetpp2023}, TartanAir V2~\cite{tartanair2020,tartanairv2}, Matterport3D~\cite{matterport3d2017}, ScanNet~\cite{scannet2017}, PointOdyssey~\cite{pointodyssey2023}, Dynamic Replica~\cite{dynamicstereo2023}, Spring~\cite{spring2023}, Virtual KITTI 2~\cite{vkitti2_2020}, DL3DV~\cite{dl3dv2024}, Waymo~\cite{waymo2020,waymo_open_dataset} and RealEstate10K~\cite{realestate10k2018}. We initialize the LoRA from the released DVD weights~\cite{dvd2026} and train it together with the prediction heads using AdamW~\cite{adamw2019}. The LoRA has rank $512$, and training takes about $40$ hours on $128$ NVIDIA GH200 GPUs.
We will release the trained model weights and code to support further research.

\vspace{3pt}\noindent\textbf{Evaluation.} At test time, input frames are resized to approximately $288\times512$ pixels, preserving their aspect ratio. All methods receive videos of consecutive frames, with no evaluation sequences included in our training data.
We evaluate video depth on Sintel~\cite{sintel2012}, Bonn~\cite{bonn2019} and KITTI~\cite{kitti2012,kittidepth2017} using absolute relative error (AbsRel) and the fraction of pixels with a depth ratio below $1.25$ ($\delta<1.25$), after one alignment per video in each method's output space.
Temporal consistency is measured by temporal alignment error (TAE)~\cite{depthanyvideo2025} on ScanNet. Camera trajectories are evaluated by absolute trajectory error (ATE) and relative pose error (RPE) on Sintel, the dynamic sequences of TUM RGB-D~\cite{tumrgbd2012} (TUM-dynamics), and ScanNet. We evaluate 3D reconstruction on 7-Scenes~\cite{sevenscenes2013} and NRGBD~\cite{nrgbd2022} using accuracy, completeness, and normal consistency, after similarity alignment and ICP.

\vspace{3pt}\noindent\textbf{Baselines.} We compare with methods built on video generation models: DepthCrafter~\cite{depthcrafter2025}, Depth Any Video~\cite{depthanyvideo2025}, DKT~\cite{dkt2025}, DVD~\cite{dvd2026}, ChronoDepth~\cite{chronodepth2025}, GeometryCrafter~\cite{geometrycrafter2025}, GenCeption~\cite{genception2026}, Aether~\cite{aether2025}, and Geo4D~\cite{geo4d2025}, each run with its official inference settings. We also compare with geometry foundation models: VGGT~\cite{vggt2025}, $\pi^3$~\cite{pi3_2026}, MapAnything~\cite{mapanything2026}, Depth Anything 3 (DA3)~\cite{depthanything3_2026}, and VGGT-$\Omega$~\cite{vggtomega2026}.

\vspace{3pt}\noindent\textbf{The supplementary material} provides implementation and evaluation details, additional ablations, more analyses of ClipShift and LayerTap, comparisons with training from random initialization and with one-frame input, and more qualitative results, including interactive point clouds and videos.

\subsection{Comparison with the State of the Art}
\label{sec::sota}

\noindent\textbf{Methods built on video generation models.}
\Cref{tab::main} compares \method with nine methods built on video generation models on the same videos.
Among them, \method achieves the lowest ATE on all three datasets, the lowest reconstruction accuracy and completeness errors on both datasets, and the most consistent video depth.
On ScanNet and TUM-dynamics, \method reduces ATE by $55.1\%$ and $56.3\%$, respectively, compared with the best of GeometryCrafter, Aether and Geo4D. Its reconstruction accuracy error on 7-Scenes falls from $0.047$ for GeometryCrafter to $0.018$, a $61.7\%$ reduction. \method also improves temporal depth consistency, reducing TAE by $30.2\%$ compared with Geo4D. Together, these results demonstrate that a video generation backbone can support accurate joint depth and camera prediction without iterative denoising or separate pose estimation.
In \cref{fig::traj}, \method follows the ground-truth camera trajectories more closely, while GeometryCrafter, Aether and Geo4D exhibit substantial drift.
The depth comparisons in \cref{fig::qualvd} further show that several baselines underestimate the depth of the person in Bonn and nearby surfaces in Sintel, whereas \method better preserves their relative positions.

\vspace{3pt}\noindent\textbf{Geometry foundation models.}
Compared with VGGT~\cite{vggt2025}, a representative image-backbone model, \method reduces video depth AbsRel by $14.2\%$ to $32.7\%$ across Sintel, Bonn and KITTI, and TAE by $13.4\%$.
It also improves camera trajectory estimation on Sintel and reduces reconstruction accuracy and completeness errors on 7-Scenes by $14.3\%$ and $29.2\%$, respectively.
Across all geometry foundation models, \method achieves the lowest depth AbsRel on Bonn and the lowest reconstruction completeness error on 7-Scenes.
While camera estimation remains less accurate than some geometry foundation models, these results demonstrate that a video generation backbone can already achieve competitive depth and reconstruction accuracy.
The reconstructions in \cref{fig::qual} illustrate this advantage in scene consistency: VGGT produces a fragmented wall on 7-Scenes, while VGGT and DA3 duplicate wall surfaces on ScanNet. In contrast, \method reconstructs these structures more coherently.

\vspace{3pt}\noindent\textbf{Inference time.}
As shown in \cref{tab::time}, \method takes $74$ to $84$\,ms per frame while predicting both depth and cameras. Among video depth estimators, only DVD is faster on Sintel and Bonn, while the others take $1.3$ to $4.2$ times as long.
On ScanNet, \method is about $4$ and $14$ times faster than Aether~\cite{aether2025} and Geo4D~\cite{geo4d2025}, respectively. Both baselines rely on iterative denoising, whereas our \method predicts depth and cameras without denoising.

\vspace{-1pt}
\subsection{Ablation}
\vspace{-1pt}
\label{sec::abl}
\begin{table}[t]\centering\footnotesize
\setlength{\tabcolsep}{5pt}
\centering
\caption{\textbf{Ablation of ClipShift and LayerTap} on eval20 and ARKit43: camera RPE-r (degrees) and reconstruction F@5 (F-score at 5\,cm).
\emph{Component ablation}: otherwise identical runs of $10{,}000$ steps on $16$ GPUs; \xmark{} and const.\ use the decoded inverse depth without a shift or with one constant shift across clips, respectively, while keeping the predicted cameras unchanged.
\emph{Final model}: trained for $30{,}000$ steps on $128$ GPUs, with reconstructions using either predicted or ground-truth (GT) cameras; `GT fit' recovers depth by fitting a scale and shift between the decoded and ground-truth inverse depth over the clip.
}
\vspace{-0.8em}
\label{tab::abl}
\begin{tabular}{@{}cccccc@{}}
\toprule
ClipShift & LayerTap &
\multicolumn{2}{c}{eval20} &
\multicolumn{2}{c}{ARKit43}\\
\cmidrule(lr){3-4}\cmidrule(lr){5-6}
shift & camera head &
RPE-r$\downarrow$ & F@5$\uparrow$ &
RPE-r$\downarrow$ & F@5$\uparrow$\\
\midrule
\multicolumn{6}{@{}l}{\emph{Component ablation (10k steps)}}\\
\xmark & \xmark & 7.80 & 19.0 & 7.78 & 22.6\\
\xmark & \cmark & \textbf{3.79} & 19.6 &
\textbf{3.41} & 21.8\\
const. & \cmark & \textbf{3.79} & 44.3 &
\textbf{3.41} & 55.2\\
\cmark & \xmark & 7.80 & 61.6 & 7.78 & 60.4\\
\cmark & \cmark & \textbf{3.79} & \textbf{65.0} &
\textbf{3.41} & \textbf{64.4}\\
\midrule
\multicolumn{6}{@{}l}{\emph{Final model (full training)}}\\
\cmark & \cmark & 1.14 & 76.8 & 1.29 & 79.3\\
\cmark & GT & -- & \textbf{85.6} & -- & 92.4\\
GT fit & GT & -- & 84.5 & -- & \textbf{93.2}\\
\bottomrule
\end{tabular}
\vspace{-2em}
\end{table}

\Cref{tab::abl} evaluates ClipShift and LayerTap in otherwise identical runs of $10{,}000$ steps, \cref{tab::dw} examines inference windows and training data, and \cref{tab::video} studies the effects of temporal input and video pretraining in a separate experiment.

\vspace{3pt}\noindent\textbf{ClipShift.}
ClipShift predicts a shared shift for each video clip to correct the geometric distortion caused by the decoded inverse depth.
Without this shift, the reconstruction F-score at 5\,cm (F@5)~\cite{eth3d2017} falls from \cmTapsFullEF{} to \cmTapsNoShiftEF{} on eval20 and from \cmTapsFullAF{} to \cmTapsNoShiftAF{} on ARKit43 (\cref{tab::abl}). A single constant shift across all clips only partially recovers this drop, reaching \cmTapsConstEF{} and \cmTapsConstAF{}, respectively.
These results show that the shift must adapt to each clip to correct the distortion in reconstructed geometry.

After full training, the final model improves both the reconstruction and the cameras (last three rows of \cref{tab::abl}): with its predicted cameras, F@5 rises from $65.0$ to $76.8$ on eval20 and from $64.4$ to $79.3$ on ARKit43, and RPE-r drops from $3.79^\circ$ to $1.14^\circ$ and from $3.41^\circ$ to $1.29^\circ$.
To assess the accuracy of the predicted shift independently of camera poses errors, we also evaluate the final model with ground-truth (GT) cameras.
ClipShift then achieves an F@5 of $85.6$ on eval20 and $92.4$ on ARKit43, within $1.1$ points of a per-clip scale-and-shift fit of the same decoded inverse depth to the ground truth (GT fit).
This shows that ClipShift, without access to the ground truth, recovers nearly all the reconstruction accuracy of the GT fit.

\vspace{3pt}\noindent\textbf{LayerTap.}
LayerTap combines features from intermediate and final transformer blocks to predict the camera of every frame.
Using only the final-block tokens increases RPE-r from \cmTapsFullERot{} to \cmNoTapsFullERot{} on eval20 and from \cmTapsFullARot{} to \cmNoTapsFullARot{} on ARKit43. Reconstruction F@5 also decreases from \cmTapsFullEF{} to \cmNoTapsFullEF{} and from \cmTapsFullAF{} to \cmNoTapsFullAF{}, respectively.
These results show that intermediate-block features improve camera estimation, which also benefits reconstruction.

\begin{table}[t]\centering\footnotesize
\setlength{\tabcolsep}{2.4pt}
\caption{\textbf{Windows and training data} on the benchmarks of \cref{tab::main}. Models are trained on three datasets, while the final model is trained longer on all eleven datasets.}
\vspace{-1em}
\label{tab::dw}
\begin{tabular}{@{}cccccccc@{}}
\toprule
 & & \multicolumn{2}{c}{AbsRel$\downarrow$} & & \multicolumn{2}{c}{ScanNet} & NRGBD\\
\cmidrule(lr){3-4}\cmidrule(lr){6-7}\cmidrule(lr){8-8}
Windows & Datasets & Sintel & Bonn & TAE$\downarrow$ & ATE$\downarrow$ & RPE-r$\downarrow$ & Acc$\downarrow$\\
\midrule
\xmark & 3 & 0.340 & 0.092 & 1.050 & 0.167 & 8.87 & 0.109\\
\cmark & 3 & 0.339 & 0.092 & 0.751 & 0.045 & 0.73 & 0.071\\
\cmark & 11 & \textbf{0.254} & \textbf{0.037} & \textbf{0.734} & \textbf{0.035} & \textbf{0.54} & \textbf{0.032}\\
\bottomrule
\end{tabular}
\vspace{-0.5em}
\end{table}

\begin{table}[t]\centering\footnotesize
\setlength{\tabcolsep}{4pt}
\caption{\textbf{Video input and pretraining} on held-out clips of TUM RGB-D, ETH3D and ARKitScenes: medians over clips of F@5, with the ground-truth cameras and one scale and shift per clip, and of the drift $G_5$.
}
\vspace{-1em}
\label{tab::video}
\begin{tabular}{@{}lcccc@{}}
\toprule
 & Video & Pretrained & & \\
Model & input & DiT & F@5$\uparrow$ & $G_5\downarrow$\\
\midrule
Wan-Scratch & \cmark & \xmark & 45.9 & 5.9\\
\vparm, one-frame videos & \xmark & \cmark & 49.2 & 18.0\\
\vparm & \cmark & \cmark & \textbf{75.3} & \textbf{5.4}\\
\bottomrule
\end{tabular}
\vspace{-1.6em}
\end{table}

\vspace{3pt}\noindent\textbf{Windows and training data.}
We examine how overlapping inference windows and additional training data affect reconstruction on long videos in \cref{tab::dw}.
For the model trained on three datasets, processing a 90-frame ScanNet video as a single input gives an RPE-r of $8.87^\circ$ and a TAE of $1.050$. Using overlapping windows reduces these errors to $0.73^\circ$ and $0.751$, respectively.
This shows the benefit of keeping inference inputs closer to the training lengths.
Training on all eleven datasets further reduces AbsRel from $0.339$ to $0.254$ on Sintel ($25.1\%$) and from $0.092$ to $0.037$ on Bonn ($59.8\%$), and reconstruction accuracy error from $0.071$ to $0.032$ on NRGBD ($54.9\%$).
These gains across both depth estimation and 3D reconstruction show the benefit of scaling up the training data and duration.

\vspace{3pt}\noindent\textbf{Video input and pretraining.}
To understand the roles of video input and generative pretraining in VG3R, we compare a Wan2.1-based depth model with frame-wise input and randomly initialized DiT variants in~\cref{tab::video}.
All models are evaluated on held-out clips from TUM RGB-D~\cite{tumrgbd2012}, ETH3D SLAM~\cite{badslam2019,eth3d2017}, and ARKitScenes~\cite{arkitscenes2021}.
We measure cross-frame drift by $G_5$, the F@5 gained by fitting one scale per frame instead of one per clip; lower values indicate more consistent depth. Wan-LoRA achieves an F@5 of $75.3$ and a $G_5$ of $5.4$.
With one-frame input, its F@5 falls to $49.2$ and $G_5$ rises to $18.0$, demonstrating the importance of temporal context for both accuracy and consistency.
Training the same DiT from random initialization (Wan-Scratch) with the same data, supervision, and steps yields an F@5 of $45.9$. These results show that temporal input is important for consistent depth prediction, while DiT pretraining provides a substantial accuracy advantage under the evaluated training budget.

\raggedbottom
\FloatBarrier
\section{Conclusion} \label{sec::conclusion}
We presented \method, which turns a pretrained video generation model into a feed-forward 3D reconstruction model. ClipShift corrects the shift ambiguity in decoded inverse depth, while LayerTap predicts per-frame cameras from transformer features. Across video depth, camera trajectory, and 3D reconstruction benchmarks,
\method achieves the lowest ATE and reconstruction accuracy and completeness errors among methods built on video generation models.
By avoiding iterative denoising, it runs about $4$ and $14$ times faster than Aether and Geo4D, respectively, while achieving competitive depth and reconstruction accuracy against geometry foundation models.

\vspace{4pt}\noindent\textbf{Future work.} Our approach could extend to other 3D and 4D perception tasks, such as surface normal estimation, scene flow, and 3D point tracking. More broadly, adapting video generators through LoRA and prediction heads raises the possibility of a unified model that both generates videos and reconstructs their underlying 3D scenes.

\section*{Acknowledgments}
This work was supported by the U.S.\ National Science Foundation under Grant No.~2316400.
This work was also supported by the Wallenberg AI, Autonomous Systems and Software Program (WASP) funded by the Knut and Alice Wallenberg Foundation.
The computations and data handling were enabled by the Arrhenius HPC resource provided by National Academic Infrastructure for Supercomputing in Sweden. 
The authors also acknowledge the South Dakota State University High Performance Computing (SDSU HPC) team for providing access to and support for the university's high-performance computing resources.

\flushbottom
\FloatBarrier

\clearpage
\maketitlesupplementary
\appendix
\makeatletter\def\@flsetnum#1{}\makeatother
\setcounter{dbltopnumber}{2}
\setcounter{dblbotnumber}{0}
\makeatletter\setlength{\@dblfptop}{0pt}\setlength{\@fptop}{0pt}\makeatother

This supplement first completes the description of \method: the details of ClipShift, LayerTap and the inference on long videos (\cref{sec::supp-impl}) and its two depth losses (\cref{sec::supp-depth}).
It then details the evaluation of the main paper (\cref{sec::supp-bench}), shows that the video depth results of \method on Sintel and Bonn hold under the evaluation code of $\pi^3$ (\cref{sec::supp-further}), and shows more reconstructions, video depth and camera trajectories (\cref{sec::supp-qual}).
The analyses that follow show that the shift predicted by ClipShift is nearly as accurate as a fit to the ground truth and that the intermediate blocks read by LayerTap halve the rotation error of the cameras in shorter runs (\cref{sec::supp-final}), and that video pretraining and temporal input contribute substantially to the accuracy of the depth (\cref{sec::supp-pretrain}).
\Cref{sec::supp-data} lists the data and their licenses.

\section{Implementation Details}
\label{sec::supp-impl}
This section gives the architecture of ClipShift and LayerTap and the inference of \method on videos longer than the training clips.
Frames are scaled to $[-1,1]$ before the VAE encoder.

\paragraph{ClipShift.}
ClipShift reads the $1536$-dimensional tokens that the last DiT block passes to the output layer, $N\times(H/16)\times(W/16)$ of them for a clip of $N$ frames of $H\times W$ pixels.
After a layer normalization and without a position embedding, one learned query attends to all of them in one multi-head attention layer with $12$ heads.
The pooled vector passes a layer normalization, a linear layer to $256$ channels, a GELU and a linear layer to one value, and a softplus with an offset of $10^{-4}$ gives the positive shift $\hat{c}$ of the clip.
The head has $9.8$M parameters, and $\tau=\epsilon=10^{-3}$ in \cref{eq::depth}.
One shift for the whole clip keeps all its frames on one affine map, as the read-out of \cref{sec::dec} assumes, whereas a shift per frame could absorb the drift between frames.

\paragraph{LayerTap.}
Each of the three token sets passes a layer normalization and a linear map before they are summed.
The head keeps all $576$ tokens of every latent frame at $288\times512$ pixels, with a 2D sinusoidal position embedding, and two learned embeddings mark latent frame $0$ and the latent frame paired with it.
Learned queries, one for latent frame $0$ and four for every later latent frame, carry an embedding of their frame index and attend jointly over each pair in four transformer blocks.
The camera head of VGGT~\cite{vggt2025} then attends across the $L$ frame tokens and refines the camera of every frame in $K=4$ iterations, as a translation, a unit quaternion and the horizontal and vertical fields of view.
In \cref{eq::cam}, $\gamma=0.6$, $w_t=3$ and $w_f=0.5$, and $\mathcal{L}_{\mathrm{pts}}$ uses the valid pixels of a grid of stride $4$.
Keeping all tokens instead of pooling them by $2\times2$ gives the lowest RPE-r on eval20 in the shorter runs, $2.75^\circ$ against $3.79^\circ$ (\cref{tab::supp-short}), so the final model keeps them.

\paragraph{Inference.}
All frames share the intrinsic matrix formed from $\hat{\bm{f}}_0$, and a longer video takes the median of $\hat{\bm{f}}_0$ over its windows.
We chose the window length and stride before any benchmark run, with an earlier model trained on three datasets, on $30$ validation clips of ScanNet~\cite{scannet2017} with $97$ frames each (\cref{tab::supp-windows}).
Windows of $37$ frames give the lowest AbsRel, and among them a stride of $24$ gives a lower ATE than a stride of $18$ with fewer passes, and a lower AbsRel and ATE than a stride of $30$, so we use a window of $37$ frames with a stride of $24$.
Processing the $97$ frames in one pass instead raises ATE from $0.090$ to $0.350$ and RPE-r from $1.02^\circ$ to $13.96^\circ$, so the windows are needed for videos longer than the training clips.

\begin{table*}[!tb]\centering\footnotesize
\setlength{\tabcolsep}{4.2pt}
\caption{\textbf{Window length and stride} on $30$ ScanNet validation clips of $97$ frames, for an earlier model trained on three datasets. Passes: inputs per clip. Bold: best among the windows.}
\label{tab::supp-windows}
\begin{tabular}{@{}cccccccc@{}}
\toprule
Window & Stride & Passes & AbsRel$\downarrow$ & ATE$\downarrow$ & RPE-t$\downarrow$ & RPE-r$\downarrow$ & TAE$\downarrow$\\
\midrule
\multicolumn{2}{c}{whole clip} & 1 & 0.0502 & 0.3498 & 0.2175 & 13.96 & 1.873\\
21 & 16 & 7 & 0.0482 & 0.0886 & \textbf{0.0296} & \textbf{0.94} & 1.431\\
29 & 24 & 5 & 0.0477 & \textbf{0.0856} & 0.0299 & 0.95 & 1.433\\
37 & 18 & 6 & \textbf{0.0454} & 0.0941 & 0.0347 & 1.07 & \textbf{1.341}\\
37 & 24 & 5 & 0.0460 & 0.0898 & 0.0323 & 1.02 & 1.419\\
37 & 30 & 4 & 0.0466 & 0.0906 & 0.0308 & 1.01 & 1.436\\
\bottomrule
\end{tabular}
\end{table*}

\section{Depth Losses}
\label{sec::supp-depth}
This section defines $\mathcal{L}_{\mathrm{dec}}$ and $\mathcal{L}_{\mathrm{mv}}$ of \cref{eq::total}, which train the depth under the alignment and the cameras of the reconstruction.
Here $\hat{\bm{x}}$ is the decoded inverse depth of \cref{eq::forward} and $x_i$ its target of \cref{eq::target}.

\subsection{Clip-Level Decoded Loss}
\label{sec::dec}
To reconstruct a scene from $\hat{\bm{x}}$, we convert all frames of a clip to depth with one affine map in inverse depth and fuse the back-projected points under one scale.
$\mathcal{L}_{\mathrm{dec}}$ fits this map to the decoded output and penalizes the remaining depth error, which contains the drift of scale and shift between frames.
It extends the decoded affine-invariant loss of E2E-FT~\cite{e2eft2025} from one image to one clip, since a fit per frame would absorb the drift.

Let $\Omega$ hold the pixels of all frames with valid depth and a target $10^{-6}<x_i(\bm{p})<1-10^{-6}$, $\Omega_i$ those of frame $i$, and let $\hat{x}$, $d$ and $\delta$ be the prediction, the depth and the inverse depth at one pixel.
A least-squares fit of $a\hat{x}+b$ to $\delta$ over $\Omega$ without gradient keeps in $\Omega'$ the pixels whose absolute residual is at most its 90th percentile, and a second fit over $\Omega'$ keeps the gradient,
\begin{equation}
\label{eq::fit}
a=\frac{mS_{x\delta}-S_{x}S_{\delta}}{mS_{xx}-S_{x}^{2}+\kappa m^{2}},\qquad
b=\frac{S_{\delta}-aS_{x}}{m},
\end{equation}
where $m=\lvert\Omega'\rvert$, $S_{x}$, $S_{\delta}$, $S_{xx}$ and $S_{x\delta}$ sum $\hat{x}$, $\delta$, $\hat{x}^2$ and $\hat{x}\delta$ over $\Omega'$, and $\kappa=10^{-8}$.
With $\hat{s}=a\hat{x}+b$ and the floor $s_{\min}=1/\max\bigl(d_{\max},1.5\,Q_{0.99}(d)\bigr)$, where $d_{\max}=20$\,m and $Q_{0.99}(d)$ is the 99th percentile of the depth over $\Omega$,
\begin{equation}
\label{eq::dec}
\begin{gathered}
r=-\ln\max(\hat{s},s_{\min})-\ln d,\qquad \mu=\operatorname*{median}_{\Omega}\,r,\\
\mathcal{L}_{\mathrm{dec}}=\tfrac{1}{5\ln2}\Bigl(\operatorname*{mean}_{\Omega}\lvert r-\mu\rvert+\operatorname*{mean}_{\Omega}\,[1-\hat{s}/s_{\min}]_{+}\Bigr),
\end{gathered}
\end{equation}
with $[\cdot]_{+}=\max(\cdot,0)$ and a stop-gradient on $\mu$.
Here $e^{-\mu}$ is the scale of the read-out, so $r-\mu$ is the log-depth error after the read-out, and the hinge penalizes depths beyond $1/s_{\min}$.
Because $a$ and $b$ keep their gradient, $\mathcal{L}_{\mathrm{dec}}$ is invariant to $\hat{\bm{x}}\mapsto\alpha\hat{\bm{x}}+\beta$ for $\alpha\neq0$ up to the ridge term, so its gradient acts only on the differences between frames and on the shape within frames, and $\mathcal{L}_{\mathrm{lat}}$ sets the global normalization.
A sample with fewer than $1000$ pixels in $\Omega$ or a variance of $\hat{x}$ below $10^{-4}$ over $\Omega'$ has no $\mathcal{L}_{\mathrm{dec}}$, and the gradient passes through the frozen decoder for every latent frame.

\subsection{Multi-View Loss}
\label{sec::mv}
$\mathcal{L}_{\mathrm{mv}}$ compares frames at correspondences given by the ground-truth depth and cameras, whereas consistent video depth~\cite{cvd2020} uses optical flow at test time.
The clip has intrinsics $\bm{K}$ and world-to-camera poses $\bm{T}_i\in SE(3)$, and $\bm{T}_j\bm{T}_i^{-1}$ has rotation $\bm{R}_{ji}$ and translation $\bm{t}_{ji}$.
The fit and the scale of \cref{eq::fit,eq::dec} give the metric depth $\tilde{D}_i(\bm{p})=e^{-\mu}/\max\bigl(a\hat{x}_i(\bm{p})+b,s_{\min}\bigr)$ of every frame.
Pixel $\bm{p}$ of frame $i$ gives the ground-truth point $\bm{Y}^*=\bm{R}_{ji}\,d_i(\bm{p})\bm{K}^{-1}\tilde{\bm{p}}+\bm{t}_{ji}$, which projects to $\bm{p}^*=\pi(\bm{Y}^*)$ in frame $j$, where $\pi(\bm{Y})$ is formed by the first two coordinates of $\bm{K}\bm{Y}/[\bm{Y}]_z$ and $[\cdot]_z$ is the third coordinate.
Of the pixels $\mathcal{S}_i$ of $\Omega_i$ on a grid of stride $2$, the visible set $\mathcal{V}_{ij}$ keeps those with $[\bm{Y}^*]_z>d_{\min}=0.05$\,m and $\bm{p}^*$ inside the image with its four neighbors in $\Omega_j$, except at depth edges, where these neighbors differ in depth by more than a factor of $1.05$, and at occlusions, where $\lvert[\bm{Y}^*]_z-d_j(\bm{p}^*)\rvert>0.03\,d_j(\bm{p}^*)$.
These correspondences come from ground truth only and carry no gradient.
The predicted point $\hat{\bm{Y}}=\bm{R}_{ji}\tilde{D}_i(\bm{p})\bm{K}^{-1}\tilde{\bm{p}}+\bm{t}_{ji}$ is compared with the depth that frame $j$ predicts at $\bm{p}^*$, interpolated bilinearly,
\begin{equation}
\label{eq::mvres}
\rho_{ij}(\bm{p})=\ln\max\bigl([\hat{\bm{Y}}]_z,\,d_{\min}\bigr)-\ln\tilde{D}_j(\bm{p}^*).
\end{equation}
The pairs $\Pi$ have $\lvert i-j\rvert\in\{1,2,4,8\}$ in both orders, and $\Pi'\subseteq\Pi$ keeps those with $\lvert\mathcal{V}_{ij}\rvert\ge\max(500,\,0.05\lvert\mathcal{S}_i\rvert)$,
\begin{equation}
\label{eq::mv}
\mathcal{L}_{\mathrm{mv}}=\frac{1}{5\ln2}\,
\frac{\sum_{(i,j)\in\Pi'}\sum_{\bm{p}\in\mathcal{V}_{ij}}\lvert\rho_{ij}(\bm{p})\rvert}{\sum_{(i,j)\in\Pi'}\lvert\mathcal{V}_{ij}\rvert},
\end{equation}
and $\mathcal{L}_{\mathrm{mv}}=0$ when $\Pi'$ is empty.
Since all frames share one map and one scale, scale drift appears directly in $\rho_{ij}$, and the gaps of $2$, $4$ and $8$ frames expose drift that is small between neighboring frames.
We apply $\mathcal{L}_{\mathrm{mv}}$ to the ScanNet++ DSLR and TartanAir clips and not to the ScanNet++ iPhone clips, whose alignment of depth and image we could not verify.
The two losses thus supervise what a per-frame loss cannot: $\mathcal{L}_{\mathrm{dec}}$ the drift of scale and shift across the frames of a clip, and $\mathcal{L}_{\mathrm{mv}}$ the agreement of the frames at the same 3D points.

\section{Evaluation Details}
\label{sec::supp-bench}
This section describes how the comparisons of \cref{tab::main} and the inference times of \cref{tab::time} are obtained.
Every method receives only the frames of a video, without cameras or ground truth, and runs once on GH200 GPUs with the settings of \cref{tab::supp-bench}.
\Cref{tab::supp-bench} lists the input size, the window and the output of every method, and the fit of its video depth: a scale and shift for an output defined up to both ($s,t$) and a scale otherwise ($s$).
For video depth, we use the 14 sequences of Sintel~\cite{sintel2012}, 5 of Bonn~\cite{bonn2019} and 13 of KITTI~\cite{kitti2012,kittidepth2017} with up to 110 frames, as prepared by MonST3R~\cite{monst3r2025}, and align each method to the ground truth by one fit per sequence in its own output space.
The temporal alignment error (TAE)~\cite{depthanyvideo2025} compares the depth of consecutive frames after reprojection with the ground-truth cameras, computed with the code of Video Depth Anything~\cite{videodepthanything2025}.
Camera trajectories are evaluated by ATE and RPE after a similarity alignment on Sintel, the 8 dynamic sequences of TUM RGB-D~\cite{tumrgbd2012} (TUM-dynamics) and the 100 test scenes of ScanNet~\cite{scannet2017}, 90 frames at stride 3 each, of which 94 have valid poses.
For reconstruction, we take the first 90 frames at stride 3 of the 18 test sequences of 7-Scenes~\cite{sevenscenes2013} and the 9 scenes of NRGBD~\cite{nrgbd2022}, back-project the predicted depth with the predicted intrinsics and cameras, align the point maps to the ground truth by a similarity transform and ICP, and report accuracy, completeness and normal consistency.

We use the official inference settings, except that Depth Any Video runs without its ensemble of 20 predictions and resizes the frames instead of cropping them, DVD uses the settings of its README and the v1.1 weights instead of the evaluation settings of its paper, GenCeption processes videos longer than 81 frames in windows instead of subsampling them, and DepthCrafter and GeometryCrafter receive the full frames without the crops of their own evaluations.
Aether predicts its cameras from center crops, so its trajectories and reconstruction come from these crops, and its reconstruction is scored on the ground truth inside them, with camera translations at the scale of its own point clouds.
GeometryCrafter predicts point maps without cameras, so its cameras come from the structure-from-motion code of its release, run without the hand-drawn masks of moving objects that no other method receives.
Apart from these exceptions, every method receives the same frames and the same alignment, so the comparisons of \cref{tab::main} reflect the predictions rather than the evaluation.
With its official settings, Aether reaches an AbsRel of 0.117 on KITTI in our runs, against 0.056 in its paper~\cite{aether2025}, while our values on Sintel and Bonn, 0.270 and 0.249, are lower than the reported 0.324 and 0.273.
Two of these three values are better in our runs than in its paper, so our setup does not systematically disadvantage Aether.
Inference times exclude model loading and start from the decoded frames in memory, except for Aether and Geo4D, whose official pipelines read the frames and include post-processing and, for Geo4D, a global alignment that takes about half of its time; reading the frames adds 15 to 23\,ms per frame to \method.
The reconstructions shown in the main paper omit the black border of the ScanNet frames for every method.

\begin{table*}[!tb]\centering\footnotesize
\setlength{\tabcolsep}{4pt}
\caption{\textbf{Settings of the compared methods}: input size $H\times W$ on Sintel / Bonn / KITTI / ScanNet, with the size of Bonn for TUM-dynamics, 7-Scenes and NRGBD, window in frames, output, and fit of the video depth to the ground truth by scale and shift ($s,t$) or by scale ($s$).}
\label{tab::supp-bench}
\begin{tabular}{@{}l>{\raggedright\arraybackslash}p{0.37\textwidth}>{\raggedright\arraybackslash}p{0.2\textwidth}lc@{}}
\toprule
Method & Input $H\times W$ & Window & Output & Fit\\
\midrule
\multicolumn{5}{@{}l}{\emph{Built on video generation models}}\\
DepthCrafter~\cite{depthcrafter2025} & $448{\times}1024$ / $512{\times}640$ / $384{\times}1280$ / $640{\times}832$ & 110, overlap 25 & disparity & $s,t$\\
Depth Any Video~\cite{depthanyvideo2025} & $416{\times}960$ / $480{\times}640$ / $288{\times}960$ / $704{\times}960$ & 32, longer by key frames and interpolation & disparity & $s,t$\\
DKT~\cite{dkt2025} & $480{\times}832$ & 81, overlap of 3 latent frames & disparity & $s,t$\\
DVD~\cite{dvd2026} & $480{\times}1136$ / $480{\times}640$ / $480{\times}1600$ / $480{\times}656$ & 81, overlap 21 & disparity & $s,t$\\
ChronoDepth~\cite{chronodepth2025} & $448{\times}1024$ / $512{\times}640$ / $320{\times}1024$ / $768{\times}1024$ & 10, overlap 5 & affine depth & $s,t$\\
GeometryCrafter~\cite{geometrycrafter2025} & $448{\times}1024$ / $512{\times}640$ / $384{\times}1216$ / $512{\times}640$ & 110, overlap 25 & point map & $s$\\
GenCeption~\cite{genception2026} & $480{\times}832$ & 81, stride 60 & depth & $s$\\
Aether~\cite{aether2025} & $480{\times}1127$ / $540{\times}720$ / $480{\times}{\sim}1590$ / $538{\times}720$ in tiles of $480{\times}720$; cameras from a $480{\times}720$ center crop & 41, stride 8; cameras: 41, stride 32 & disparity, cameras & $s$\\
Geo4D~\cite{geo4d2025} & $256{\times}576$ / $384{\times}512$ / $192{\times}640$ / $384{\times}512$ & 16, stride 4, global alignment & depth, cameras & $s$\\
\midrule
\multicolumn{5}{@{}l}{\emph{Geometry foundation models}}\\
VGGT~\cite{vggt2025} & $224{\times}518$ / $392{\times}518$ / $154{\times}518$ / $392{\times}518$ & whole video & depth, cameras & $s$\\
$\pi^3$~\cite{pi3_2026} & $224{\times}504$ / $378{\times}504$ / $154{\times}504$ / $378{\times}504$ & whole video & point map, cameras & $s$\\
MapAnything~\cite{mapanything2026} & $336{\times}798$ / $448{\times}602$ / $280{\times}938$ / $448{\times}602$ & whole video & depth, cameras & $s$\\
DA3~\cite{depthanything3_2026} & $210{\times}504$ / $378{\times}504$ / $154{\times}504$ / $378{\times}504$ & whole video & depth, cameras & $s$\\
VGGT-$\Omega$~\cite{vggtomega2026} & $272{\times}640$ / $368{\times}480$ / $224{\times}752$ / $352{\times}480$ & whole video & depth, cameras & $s$\\
\midrule
\textbf{\method} & $256{\times}592$ / $336{\times}448$ / $208{\times}704$ / $336{\times}448$ & 37, stride 24 & depth, cameras & $s$\\
\bottomrule
\end{tabular}
\end{table*}

\section{Video Depth with the Evaluation Code of $\pi^3$}
\label{sec::supp-further}
To check that the video depth results do not depend on our alignment, \cref{tab::supp-vdepth} evaluates \method and the geometry foundation models with the video depth evaluation code of $\pi^3$~\cite{pi3_2026}, run unchanged.
The code aligns every output as depth, by a scale and shift that minimize the L1 error or by a scale alone, and its scale-only rows are the values of the main paper.
The main paper aligns each method in its own output space instead, because the inverse of an affine-invariant inverse depth is not an affine-invariant depth.
Under this code, \method still has the lowest AbsRel and the highest $\delta_{1.25}$ on Sintel and Bonn after the scale-and-shift alignment, and the highest $\delta_{1.25}$ on both after the scale alignment, so its results on these benchmarks do not depend on aligning each method in its own output space.

\begin{table*}[t]\centering\footnotesize
\setlength{\tabcolsep}{5pt}
\caption{\textbf{Video depth with the evaluation code of $\pi^3$}~\cite{pi3_2026}: AbsRel and $\delta<1.25$ after one alignment per video by scale and shift or by scale alone, and TAE~\cite{depthanyvideo2025} on 94 ScanNet sequences. Best and second best are shaded.}
\label{tab::supp-vdepth}
\begin{tabular}{@{}llccccccc@{}}
\toprule
 & & \multicolumn{2}{c}{Sintel} & \multicolumn{2}{c}{Bonn} & \multicolumn{2}{c}{KITTI} & ScanNet\\
\cmidrule(lr){3-4}\cmidrule(lr){5-6}\cmidrule(lr){7-8}\cmidrule(lr){9-9}
Alignment & Method & AbsRel$\downarrow$ & $\delta_{1.25}\uparrow$ & AbsRel$\downarrow$ & $\delta_{1.25}\uparrow$ & AbsRel$\downarrow$ & $\delta_{1.25}\uparrow$ & TAE$\downarrow$\\
\midrule
\multirow{6}{*}{Scale \& shift} & VGGT~\cite{vggt2025} & 0.227 & 0.684 & 0.049 & 0.974 & 0.059 & 0.961 & 0.848\\
 & $\pi^3$~\cite{pi3_2026} & 0.211 & \cellcolor{second}0.728 & \cellcolor{second}0.045 & \cellcolor{second}0.976 & \cellcolor{best}\textbf{0.037} & \cellcolor{second}0.986 & \cellcolor{best}\textbf{0.626}\\
 & MapAnything~\cite{mapanything2026} & 0.253 & 0.632 & 0.054 & 0.975 & 0.070 & 0.973 & 1.095\\
 & DA3~\cite{depthanything3_2026} & 0.218 & 0.683 & 0.046 & 0.973 & 0.051 & 0.977 & 1.058\\
 & VGGT-$\Omega$~\cite{vggtomega2026} & \cellcolor{second}0.194 & 0.649 & 0.115 & 0.932 & \cellcolor{second}0.039 & \cellcolor{best}\textbf{0.987} & 0.748\\
 & \textbf{\method} & \cellcolor{best}\textbf{0.182} & \cellcolor{best}\textbf{0.752} & \cellcolor{best}\textbf{0.036} & \cellcolor{best}\textbf{0.979} & 0.044 & 0.977 & \cellcolor{second}0.734\\
\midrule
\multirow{6}{*}{Scale} & VGGT~\cite{vggt2025} & 0.296 & 0.647 & 0.055 & 0.971 & 0.072 & 0.965 & \\
 & $\pi^3$~\cite{pi3_2026} & \cellcolor{best}\textbf{0.232} & 0.665 & 0.050 & \cellcolor{second}0.975 & \cellcolor{best}\textbf{0.039} & \cellcolor{second}0.986 & \\
 & MapAnything~\cite{mapanything2026} & 0.337 & 0.580 & 0.063 & 0.972 & 0.074 & 0.972 & \\
 & DA3~\cite{depthanything3_2026} & 0.313 & \cellcolor{second}0.668 & \cellcolor{second}0.049 & 0.973 & 0.057 & 0.978 & \\
 & VGGT-$\Omega$~\cite{vggtomega2026} & 0.256 & 0.660 & 0.114 & 0.933 & \cellcolor{second}0.040 & \cellcolor{best}\textbf{0.987} & \\
 & \textbf{\method} & \cellcolor{second}0.254 & \cellcolor{best}\textbf{0.699} & \cellcolor{best}\textbf{0.037} & \cellcolor{best}\textbf{0.979} & 0.058 & 0.979 & \\
\bottomrule
\end{tabular}
\end{table*}

\section{Further Qualitative Results}
\label{sec::supp-qual}
This section shows more reconstructions, video depth and camera trajectories in the format of \cref{fig::qual,fig::qualvd,fig::traj} of the main paper.

\begin{figure*}[!tb]
\centering
\includegraphics[width=\textwidth]{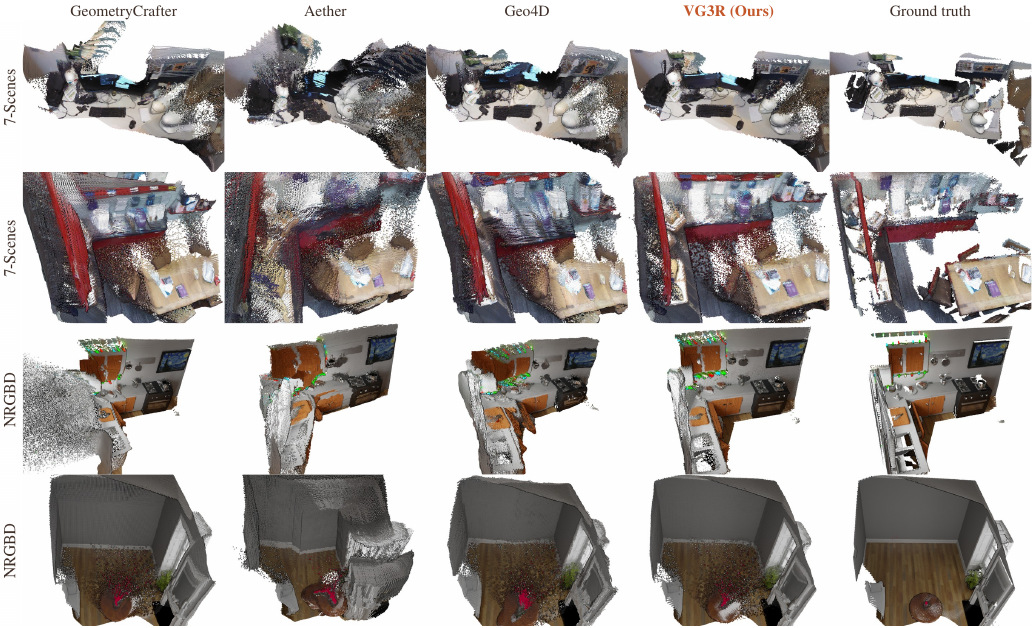}
\caption{\textbf{Reconstructions of the methods built on video generation models} on 7-Scenes heads-seq-01 and redkitchen-seq-12 and on NRGBD morning\_apartment and grey\_white\_room, in the format of \cref{fig::qual}: every method fuses the point maps of all frames with its own predicted cameras, without confidence filtering or other post-processing, and is aligned to the ground truth by a similarity transform.}
\label{fig::supp-vgrec}
\end{figure*}

\begin{figure*}[!tb]
\centering
\includegraphics[width=\textwidth]{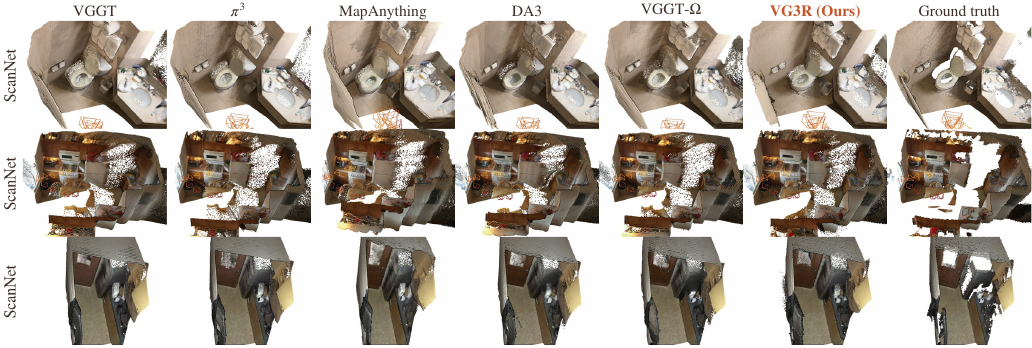}
\caption{\textbf{Reconstructions and cameras on ScanNet} scene0735\_00, scene0749\_00 and scene0743\_00, in the format of \cref{fig::qual}, with the predicted camera of every tenth frame colored by time.}
\label{fig::supp-gfmrec}
\end{figure*}

\paragraph{Reconstruction.}
\Cref{fig::supp-vgrec} compares \method with GeometryCrafter~\cite{geometrycrafter2025}, Aether~\cite{aether2025} and Geo4D~\cite{geo4d2025}, the methods built on video generation models that also predict cameras.
GeometryCrafter leaves clouds of points in front of the surfaces, Aether bends walls and furniture, and Geo4D bends the counter of the NRGBD kitchen, whereas \method keeps the layout of each room close to the ground truth.
This matches \cref{tab::main}, where the accuracy error of \method is 0.018\,m on 7-Scenes and 0.032\,m on NRGBD, against at least 0.047\,m and 0.065\,m for these three methods.
\Cref{fig::supp-gfmrec} compares \method with the geometry foundation models on further ScanNet videos.
Its point clouds and cameras are close to theirs, as its competitive accuracy in \cref{tab::main} indicates, although \method reconstructs with a video generation model instead of an image backbone.

\begin{figure*}[!tb]
\centering
\includegraphics[width=\textwidth]{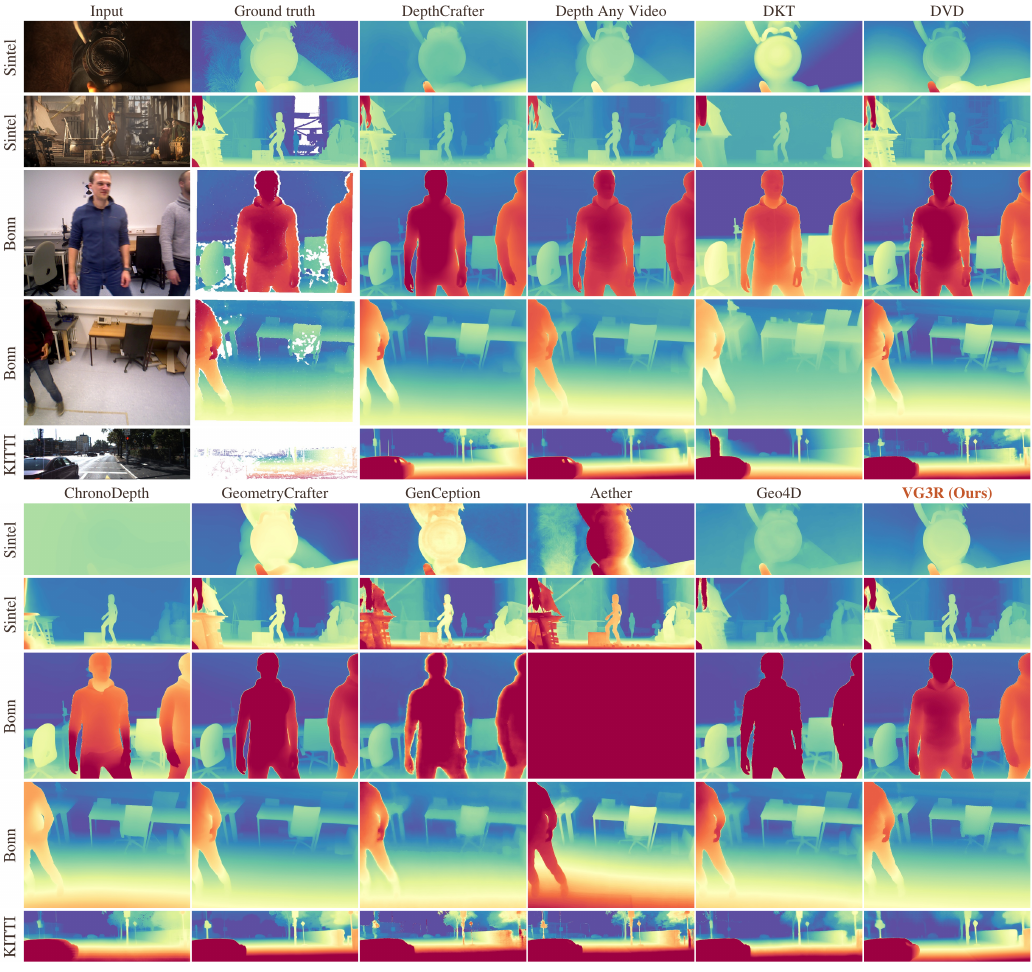}
\caption{\textbf{Video depth on further frames} of Sintel shaman\_3 and market\_2, Bonn synchronous and balloon2, and KITTI 2011\_09\_29\_drive\_0026, in the format of \cref{fig::qualvd}: inverse depth after the alignment of \cref{tab::main}, with one color range per sequence from the ground truth.}
\label{fig::supp-vdepth}
\end{figure*}

\paragraph{Video depth.}
\Cref{fig::supp-vdepth} shows the video depth of the methods built on video generation models on further frames of Sintel, Bonn and KITTI.
\method separates people, furniture and the background as the ground truth does, while ChronoDepth and DKT lose contrast and Aether fails on the first Bonn frame.
On Bonn, the AbsRel of \method in \cref{tab::main} is 0.037, against at least 0.055 for these methods.

\begin{figure*}[!tb]
\centering
\includegraphics[width=\textwidth]{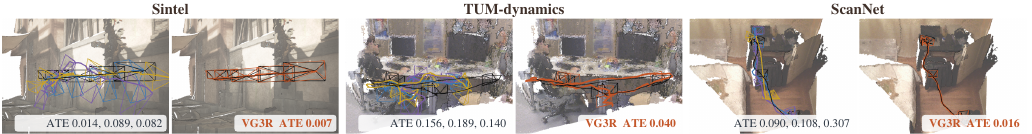}\\[2pt]
\includegraphics[width=\textwidth]{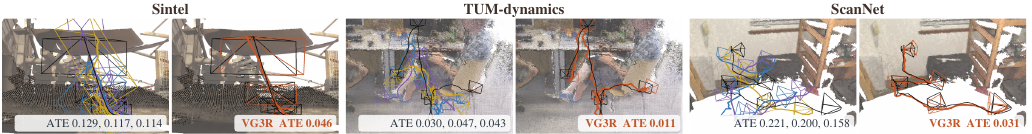}
\caption{\textbf{Camera trajectories on further videos}, in the format of \cref{fig::traj}: two more sequences per benchmark: Sintel alley\_2 and market\_6, two sequences with recognizable scenes on which \method has the lowest ATE, and TUM-dynamics walking\_rpy and sitting\_rpy and ScanNet scene0720\_00 and scene0717\_00, among the five where \method leads most in ATE. Left of each pair: GeometryCrafter (blue), Aether (purple) and Geo4D (yellow), with ATE reported in this order. Right: \method.}
\label{fig::supp-traj}
\end{figure*}

\paragraph{Camera trajectories.}
\Cref{fig::supp-traj} shows the camera trajectories of six further videos.
On each of them, the lowest ATE of GeometryCrafter, Aether and Geo4D is 2.7 to 5.6 times that of \method, whose trajectory follows the ground truth where theirs drift.

\section{Analyses of ClipShift and LayerTap}
\label{sec::supp-final}
To assess ClipShift on the final model, \cref{tab::supp-shift} reads it with the ground-truth cameras in the ways of \cref{tab::abl}.
The decoded inverse depth without a shift reaches an F@5 of \ablbNoShiftE{} on eval20 and \ablbNoShiftA{} on ARKit43, one constant shift, the mean of the predicted shifts over each set, reaches \ablbConstE{} and \ablbConstA{}, and the predicted shift, which ranges from \shiftMin{} to \shiftMax{} across these clips, reaches \ablbShiftE{} and \ablbShiftA{}, within \ablbShiftMax{} points of an affine fit to the ground truth.
The predicted cameras lower F@5 by \ablbCamE{} and \ablbCamA{} points, so the cameras, not the shift, limit the reconstruction.

\begin{table}[!tb]\centering\footnotesize
\setlength{\tabcolsep}{4pt}
\caption{\textbf{Read-out of the final model}: F@5 of the fused point cloud after one scale per clip, for different shifts of the decoded inverse depth and with ground-truth (GT) or its own cameras.}
\label{tab::supp-shift}
\begin{tabular}{@{}lcc@{}}
\toprule
Depth and cameras & eval20 F@5$\uparrow$ & ARKit43 F@5$\uparrow$\\
\midrule
No shift, GT cam. & 19.6 & 25.4\\
Constant shift, GT cam. & 53.2 & 75.2\\
Predicted shift, GT cam. & 85.6 & 92.4\\
Affine fit, GT cam. & 84.5 & 93.2\\
Predicted shift, own cam. & 76.8 & 79.3\\
\bottomrule
\end{tabular}
\end{table}

\begin{figure}[!tb]
\centering
\includegraphics[width=\linewidth]{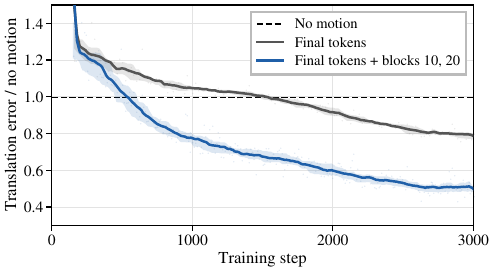}
\caption{\textbf{Inputs of the camera head over training}: translation error relative to predicting no motion, in two runs of $3000$ steps on $16$ GPUs on the same samples whose camera heads read the final tokens alone or also blocks 10 and 20, as a moving average over 300 steps with the 25th to 75th percentiles as the band.}
\label{fig::camhead}
\end{figure}

\Cref{tab::supp-short} varies the camera head, the camera loss and the precision of the heads in runs of $10{,}000$ steps on $16$ GPUs, each started from the LoRA of DVD v1.1~\cite{dvd2026} with one seed and evaluated on eval20 and ARKit43.
The base run trains on ScanNet++, TartanAir V2 and Matterport3D at $288\times512$ pixels, its camera head average-pools the tokens of each latent frame by $2\times2$ and reads blocks 10 and 20 besides the final tokens, and both heads run under bf16 autocast.
The camera head may instead read all $18\times32$ tokens, add tokens of each RGB frame, also see the previous latent frame, or be replaced by a head on the camera and register tokens of VGGT-$\Omega$~\cite{vggtomega2026} passed through the DiT, and the camera loss may be weighted twice or extended by the pairwise pose loss of VGGT-$\Omega$.
These are single short runs, and among the runs that change only the camera head, its loss or the precision of the heads, the AbsRel on eval20 spans 0.052 to 0.058, so we do not read smaller differences as effects.
Among the inputs of the camera head, the intermediate blocks matter most: without them, RPE-r doubles, from $3.79^\circ$ to $7.80^\circ$ on eval20 and from $3.41^\circ$ to $7.78^\circ$ on ARKit43, while every other change of the head, its loss or its precision moves it by at most about $1^\circ$.
Early in training, the camera head that also reads blocks 10 and 20 learns the translation faster (\cref{fig::camhead}).
Between steps 1000 and 1500, its translation error is 0.72 of that of predicting no motion, while the head on the final tokens alone, at 1.03, still errs more than predicting no motion; between steps 2400 and 3000, the two reach 0.51 and 0.81.

\begin{table*}[!tb]\centering\footnotesize
\setlength{\tabcolsep}{5pt}
\caption{\textbf{Ablations in shorter runs} of $10{,}000$ steps on $16$ GPUs: AUC@5 and RPE-r of the predicted cameras, AbsRel of the depth and F@5 of the point cloud fused with the predicted cameras, after one scale per clip. Best and second best are shaded.}
\label{tab::supp-short}
\begin{tabular}{@{}lcccccccc@{}}
\toprule
 & \multicolumn{4}{c}{eval20} & \multicolumn{4}{c}{ARKit43}\\
\cmidrule(lr){2-5}\cmidrule(lr){6-9}
Run & AUC@5$\uparrow$ & RPE-r$\downarrow$ & AbsRel$\downarrow$ & F@5$\uparrow$ & AUC@5$\uparrow$ & RPE-r$\downarrow$ & AbsRel$\downarrow$ & F@5$\uparrow$\\
\midrule
Base & 5.53 & 3.79 & \cellcolor{second}0.053 & 65.0 & 4.42 & 3.41 & \cellcolor{second}0.042 & 64.4\\
\midrule
\multicolumn{9}{@{}l}{\emph{Camera head}}\\
Final tokens only & 2.15 & 7.80 & 0.055 & 61.6 & 1.39 & 7.78 & 0.043 & 60.4\\
No pooling of the tokens & \cellcolor{second}6.15 & \cellcolor{best}\textbf{2.75} & \cellcolor{best}\textbf{0.052} & 65.7 & \cellcolor{best}\textbf{5.97} & \cellcolor{second}2.40 & \cellcolor{second}0.042 & 64.9\\
Tokens of each RGB frame & 5.60 & 4.37 & 0.058 & 63.7 & 4.50 & 3.44 & \cellcolor{second}0.042 & 65.0\\
Also the previous latent frame & 5.93 & 2.99 & 0.057 & 64.7 & 5.58 & 2.85 & \cellcolor{second}0.042 & \cellcolor{second}65.6\\
Camera and register tokens & 5.93 & 3.19 & 0.055 & \cellcolor{second}66.2 & 5.09 & \cellcolor{best}\textbf{2.31} & \cellcolor{second}0.042 & 62.3\\
\midrule
\multicolumn{9}{@{}l}{\emph{Camera loss}}\\
Weight $\times 2$ & 6.09 & 3.12 & 0.054 & \cellcolor{best}\textbf{67.8} & \cellcolor{second}5.89 & 2.91 & \cellcolor{second}0.042 & \cellcolor{best}\textbf{65.8}\\
Pairwise pose loss & 5.67 & \cellcolor{second}2.76 & 0.056 & 66.0 & 5.20 & 2.94 & 0.043 & 65.3\\
\midrule
\multicolumn{9}{@{}l}{\emph{Precision of the heads}}\\
32-bit & 5.59 & 4.12 & 0.058 & 63.2 & 4.68 & 3.52 & \cellcolor{second}0.042 & 64.2\\
32-bit, pairwise pose loss & \cellcolor{best}\textbf{6.41} & 3.34 & 0.057 & 64.4 & 4.59 & 3.16 & \cellcolor{best}\textbf{0.041} & 64.6\\
\bottomrule
\end{tabular}
\end{table*}

The final model reads all tokens of every latent frame and runs both heads in 32 bits; it trains at the four image sizes of the benchmarks, reads clips of $49$ frames in a quarter of the steps and takes $60\%$ of its microbatches from ScanNet~\cite{scannet2017}, PointOdyssey~\cite{pointodyssey2023}, Spring~\cite{spring2023}, Virtual KITTI 2~\cite{vkitti2_2020}, Dynamic Replica~\cite{dynamicstereo2023}, DL3DV-10K~\cite{dl3dv2024}, the front camera of the Waymo Open Dataset~\cite{waymo2020} and RealEstate10K~\cite{realestate10k2018}, of which DL3DV-10K and Waymo supervise only the cameras and sparse points and RealEstate10K only the cameras.
From step $15{,}000$ on, it draws the iPhone videos of ScanNet++ from a larger set of scenes at the same share of the microbatches.

\section{Video Pretraining and Temporal Input}
\label{sec::supp-pretrain}
To measure what the video pretraining and the temporal input contribute, we train depth models that share the training data, the depth supervision and the number of steps and differ only in their initialization or their input, and we evaluate them on held-out clips with the ground-truth cameras.
\vparm is Wan2.1~\cite{wan2025} with LoRA as in \method, predicting depth only.
Wan-Scratch is the DiT of Wan2.1 trained from random initialization, with the pretrained VAE.
\vparm-DVD starts from the LoRA of DVD v1.1~\cite{dvd2026}.
With one-frame input, \vparm receives every frame as a one-frame video.

\begin{table*}[!tb]\centering\footnotesize
\setlength{\tabcolsep}{4pt}
\caption{\textbf{Video pretraining, temporal input and initialization}: median of the per-clip differences on the $79$ held-out clips between the first and the second model of \cref{sec::supp-pretrain}, positive in favor of the first (a higher F@5 or a lower $G_5$), and its 90\% bootstrap interval. Solved shift: the shift of the decoded inverse depth solved from the known cameras, as for deployment; released DVD: the better of the released DVD v1.0 and v1.1 on each set.}
\label{tab::supp-pretrain}
\begin{tabular}{@{}lllccc@{}}
\toprule
Comparison & First model & Second model & Metric & Median & 90\% interval\\
\midrule
Pretraining & \vparm & Wan-Scratch & F@5 & $+22.6$ & $[+22.0, +27.1]$\\
Video input & \vparm & \vparm, one-frame input & F@5 & $+20.0$ & $[+16.5, +22.9]$\\
 &  &  & $G_5$ & $+10.0$ & $[+6.7, +12.7]$\\
\midrule
DVD start & \vparm-DVD & \vparm & F@5 & $+10.5$ & $[+8.3, +11.8]$\\
Deployment & \vparm-DVD, solved shift & released DVD, solved shift & F@5 & $+8.7$ & $[+7.0, +11.3]$\\
\bottomrule
\end{tabular}
\end{table*}

\paragraph{Training.}
All models read the same stream of microbatches and use the four depth terms of \cref{eq::total} with the same weights.
Video clips come from the DSLR and iPhone captures of ScanNet++~\cite{scannetpp2023} and from TartanAir V2~\cite{tartanair2020,tartanairv2}, and single images from ScanNet++ and Matterport3D~\cite{matterport3d2017}.
A microbatch is one clip of 5 to 37 frames or eight images, and a step averages eight microbatches.
Each model trains for $30{,}000$ steps on four GH200 GPUs with AdamW~\cite{adamw2019}, at a learning rate of $10^{-4}$ for \vparm and Wan-Scratch and of $3\times10^{-5}$ for \vparm-DVD.
The checkpoint is selected on six held-out ScanNet++ scenes.
Wan-Scratch was still improving at the end of training, with its selection score rising by 10.8, 10.1 and 5.4 points over the three blocks of $10{,}000$ steps, while that of \vparm changed by $-1.1$ points in the last block.

\paragraph{Evaluation.}
The held-out set has 79 clips of 37 frames at $288\times512$ pixels, 12 from TUM RGB-D~\cite{tumrgbd2012}, 20 from the SLAM benchmark of ETH3D~\cite{badslam2019,eth3d2017} and 47 from ARKitScenes~\cite{arkitscenes2021}.
The training data contain none of these datasets, and the whole training fold of ARKitScenes, which holds the 47 clips, is excluded from all our training, while the undisclosed pretraining data of Wan2.1 may contain their public videos.
Every model receives the RGB frames alone.
Its depth is back-projected with the ground-truth cameras and fused with one scale per clip, the median ratio of ground truth to prediction, after one affine map per clip for a model that predicts affine-invariant inverse depth.
F@5 is the F-score at 5\,cm~\cite{eth3d2017} against a mesh of the depth rendered from the laser scan for ARKitScenes and a TSDF fusion of the sensor depth for TUM RGB-D and ETH3D, and the drift $G_5$ is the F@5 gained by fitting one scale per frame instead of one per clip.
Tables report medians over clips, and a comparison of two models uses the median of their per-clip differences with a 90\% bootstrap interval.

\paragraph{Decision rules.}
Each model was evaluated once on these clips with the same evaluation code.
A model wins a comparison of F@5 when the median difference is at least 3 points in its favor, it is better on at least 70\% of the clips, and P@5 and F@10 agree.
A comparison of $G_5$ passes under the same conditions on $G_5$ if, in addition, $G_{10}$ and the spread of the per-frame scales agree and the F@5 of the first model is at most 3 points lower.
We set the margin of 3 points on development data, without reading any held-out clip.

\paragraph{Results.}
\Cref{tab::supp-pretrain} gives the comparisons, and \cref{tab::video} of the main paper the medians of \vparm, its use with one-frame input and Wan-Scratch.
The first model wins every comparison of F@5, and \vparm with video input passes the comparison of $G_5$ against its use with one-frame input.
\vparm is far more accurate than Wan-Scratch, which was still improving at the end of training.
With one-frame input, \vparm is less accurate and drifts more, as in the main paper.
\vparm-DVD, which also trains at a lower learning rate, is more accurate than \vparm.
For deployment, the shift is solved from the known cameras instead of fitted to the ground truth, by modeling depth as $s/(\hat{x}+c)$ with one $s$ and one $c$ per clip; with this shift, \vparm-DVD is more accurate than the best released DVD model.
In sum, \vparm is more accurate than Wan-Scratch after the same number of steps by a median of $22.6$ points of F@5 and than its use with one-frame input by $20.0$ points, and starting from the LoRA of DVD adds $10.5$ points, so video pretraining and temporal input both contribute substantially to the accuracy within our training budget.

\section{Data and Licenses}
\label{sec::supp-data}
ScanNet++~\cite{scannetpp2023} and Matterport3D~\cite{matterport3d2017} are used under their terms of use, which do not permit redistribution, so we redistribute no image, depth map or mesh of either dataset.
ScanNet v2~\cite{scannet2017} is used under its Terms of Use for the benchmarks, the choice of the windows and the training of the final model, and ARKitScenes~\cite{arkitscenes2021} under the license of Apple for non-commercial research.
TUM RGB-D~\cite{tumrgbd2012}, TartanAir V2~\cite{tartanairv2} and Spring~\cite{spring2023} are available under CC BY 4.0, the ETH3D SLAM sequences~\cite{badslam2019} and PointOdyssey~\cite{pointodyssey2023} under CC BY-NC-SA 4.0, Virtual KITTI 2~\cite{vkitti2_2020} under CC BY-NC-SA 3.0, and the released DVD weights~\cite{dvd2026} under CC BY-NC 4.0.
DL3DV-10K~\cite{dl3dv2024} is used under its terms of use and CC BY-NC 4.0, and Dynamic Replica~\cite{dynamicstereo2023} under its license agreement for non-commercial research.
The camera poses of RealEstate10K~\cite{realestate10k2018} are available under CC BY 4.0, while its frames are third-party videos that no license of the dataset covers, and we redistribute none of them.
The Waymo Open Dataset~\cite{waymo2020,waymo_open_dataset} is used under the Waymo Dataset License Agreement for Non-Commercial Use.


{\small
\begin{thebibliography}{91}
\providecommand{\natexlab}[1]{#1}
\providecommand{\url}[1]{\texttt{#1}}
\expandafter\ifx\csname urlstyle\endcsname\relax
  \providecommand{\doi}[1]{doi: #1}\else
  \providecommand{\doi}{doi: \begingroup \urlstyle{rm}\Url}\fi

\bibitem[way(2019-2025)]{waymo_open_dataset}
{Waymo Open Dataset: An autonomous driving dataset}.
\newblock \url{https://www.waymo.com/open}, 2019-2025.

\bibitem[Azinovi{\'c} et~al.(2022)Azinovi{\'c}, Martin-Brualla, Goldman,
  Nie{\ss}ner, and Thies]{nrgbd2022}
Dejan Azinovi{\'c}, Ricardo Martin-Brualla, Dan~B. Goldman, Matthias
  Nie{\ss}ner, and Justus Thies.
\newblock {Neural {RGB-D} Surface Reconstruction}.
\newblock In \emph{CVPR}, 2022.

\bibitem[Bahmani et~al.(2026)Bahmani, Shen, Ren, Huang, Jiang, Turki,
  Tagliasacchi, Lindell, Gojcic, Fidler, Ling, Gao, and Ren]{lyra2026}
Sherwin Bahmani, Tianchang Shen, Jiawei Ren, Jiahui Huang, Yifeng Jiang,
  Haithem Turki, Andrea Tagliasacchi, David~B. Lindell, Zan Gojcic, Sanja
  Fidler, Huan Ling, Jun Gao, and Xuanchi Ren.
\newblock {{Lyra}: Generative {3D} Scene Reconstruction via Video Diffusion
  Model Self-Distillation}.
\newblock In \emph{ICLR}, 2026.

\bibitem[Baruch et~al.(2021)Baruch, Chen, Dehghan, Dimry, Feigin, Fu, Gebauer,
  Joffe, Kurz, Schwartz, and Shulman]{arkitscenes2021}
Gilad Baruch, Zhuoyuan Chen, Afshin Dehghan, Tal Dimry, Yuri Feigin, Peter Fu,
  Thomas Gebauer, Brandon Joffe, Daniel Kurz, Arik Schwartz, and Elad Shulman.
\newblock {{ARKitScenes}: A Diverse Real-World Dataset For {3D} Indoor Scene
  Understanding Using Mobile {RGB-D} Data}.
\newblock In \emph{NeurIPS Datasets and Benchmarks}, 2021.

\bibitem[Blattmann et~al.(2023)Blattmann, Dockhorn, Kulal, Mendelevitch,
  Kilian, Lorenz, Levi, English, Voleti, Letts, Jampani, and Rombach]{svd2023}
Andreas Blattmann, Tim Dockhorn, Sumith Kulal, Daniel Mendelevitch, Maciej
  Kilian, Dominik Lorenz, Yam Levi, Zion English, Vikram Voleti, Adam Letts,
  Varun Jampani, and Robin Rombach.
\newblock {{Stable Video Diffusion}: Scaling Latent Video Diffusion Models to
  Large Datasets}.
\newblock \emph{arXiv preprint arXiv:2311.15127}, 2023.

\bibitem[Butler et~al.(2012)Butler, Wulff, Stanley, and Black]{sintel2012}
Daniel~J. Butler, Jonas Wulff, Garrett~B. Stanley, and Michael~J. Black.
\newblock {A Naturalistic Open Source Movie for Optical Flow Evaluation}.
\newblock In \emph{ECCV}, 2012.

\bibitem[Cabon et~al.(2020)Cabon, Murray, and Humenberger]{vkitti2_2020}
Yohann Cabon, Naila Murray, and Martin Humenberger.
\newblock {Virtual {KITTI} 2}.
\newblock \emph{arXiv preprint arXiv:2001.10773}, 2020.

\bibitem[Carion et~al.(2026)Carion, Gustafson, Hu, Debnath, Hu, Suris, Ryali,
  Alwala, Khedr, Huang, Lei, Ma, Guo, Kalla, Marks, Greer, Wang, Sun,
  R{\"a}dle, Afouras, Mavroudi, Xu, Wu, Zhou, Momeni, Hazra, Ding, Vaze,
  Porcher, Li, Li, Kamath, Cheng, Doll{\'a}r, Ravi, Saenko, Zhang, and
  Feichtenhofer]{sam3_2025}
Nicolas Carion, Laura Gustafson, Yuan-Ting Hu, Shoubhik Debnath, Ronghang Hu,
  Didac Suris, Chaitanya Ryali, Kalyan~Vasudev Alwala, Haitham Khedr, Andrew
  Huang, Jie Lei, Tengyu Ma, Baishan Guo, Arpit Kalla, Markus Marks, Joseph
  Greer, Meng Wang, Peize Sun, Roman R{\"a}dle, Triantafyllos Afouras,
  Effrosyni Mavroudi, Katherine Xu, Tsung-Han Wu, Yu Zhou, Liliane Momeni,
  Rishi Hazra, Shuangrui Ding, Sagar Vaze, Francois Porcher, Feng Li, Siyuan
  Li, Aishwarya Kamath, Ho~Kei Cheng, Piotr Doll{\'a}r, Nikhila Ravi, Kate
  Saenko, Pengchuan Zhang, and Christoph Feichtenhofer.
\newblock {{SAM} 3: {Segment Anything} with Concepts}.
\newblock In \emph{ICLR}, 2026.

\bibitem[Chang et~al.(2017)Chang, Dai, Funkhouser, Halber, Nie{\ss}ner, Savva,
  Song, Zeng, and Zhang]{matterport3d2017}
Angel Chang, Angela Dai, Thomas Funkhouser, Maciej Halber, Matthias
  Nie{\ss}ner, Manolis Savva, Shuran Song, Andy Zeng, and Yinda Zhang.
\newblock {{Matterport3D}: Learning from {RGB-D} Data in Indoor Environments}.
\newblock In \emph{3DV}, 2017.

\bibitem[Chen et~al.(2025)Chen, Guo, Zhu, Zhang, Huang, Feng, and
  Kang]{videodepthanything2025}
Sili Chen, Hengkai Guo, Shengnan Zhu, Feihu Zhang, Zilong Huang, Jiashi Feng,
  and Bingyi Kang.
\newblock {{Video Depth Anything}: Consistent Depth Estimation for Super-Long
  Videos}.
\newblock In \emph{CVPR}, 2025.

\bibitem[Dai et~al.(2017)Dai, Chang, Savva, Halber, Funkhouser, and
  Nie{\ss}ner]{scannet2017}
Angela Dai, Angel~X. Chang, Manolis Savva, Maciej Halber, Thomas Funkhouser,
  and Matthias Nie{\ss}ner.
\newblock {{ScanNet}: Richly-Annotated {3D} Reconstructions of Indoor Scenes}.
\newblock In \emph{CVPR}, 2017.

\bibitem[El~Banani et~al.(2024)El~Banani, Raj, Maninis, Kar, Li, Rubinstein,
  Sun, Guibas, Johnson, and Jampani]{probe3d2024}
Mohamed El~Banani, Amit Raj, Kevis-Kokitsi Maninis, Abhishek Kar, Yuanzhen Li,
  Michael Rubinstein, Deqing Sun, Leonidas Guibas, Justin Johnson, and Varun
  Jampani.
\newblock {Probing the {3D} Awareness of Visual Foundation Models}.
\newblock In \emph{CVPR}, 2024.

\bibitem[Fu et~al.(2024)Fu, Yin, Hu, Wang, Ma, Tan, Shen, Lin, and
  Long]{geowizard2024}
Xiao Fu, Wei Yin, Mu Hu, Kaixuan Wang, Yuexin Ma, Ping Tan, Shaojie Shen, Dahua
  Lin, and Xiaoxiao Long.
\newblock {{GeoWizard}: Unleashing the Diffusion Priors for {3D} Geometry
  Estimation from a Single Image}.
\newblock In \emph{ECCV}, 2024.

\bibitem[Gabeur et~al.(2026)Gabeur, Long, Peng, Voigtlaender, Sun, Bao, Truong,
  Wang, Zhou, Barron, Genova, Kannen, Ben, Li, Guo, Yogin, Gu, Chen, Wang, Xie,
  Zhou, He, Funkhouser, Alayrac, and Soricut]{gabeur2026image}
Valentin Gabeur, Shangbang Long, Songyou Peng, Paul Voigtlaender, Shuyang Sun,
  Yanan Bao, Karen Truong, Zhicheng Wang, Wenlei Zhou, Jonathan~T. Barron, Kyle
  Genova, Nithish Kannen, Sherry Ben, Yandong Li, Mandy Guo, Suhas Yogin,
  Yiming Gu, Huizhong Chen, Oliver Wang, Saining Xie, Howard Zhou, Kaiming He,
  Thomas Funkhouser, Jean-Baptiste Alayrac, and Radu Soricut.
\newblock {Image Generators are Generalist Vision Learners}.
\newblock \emph{arXiv preprint arXiv:2604.20329}, 2026.

\bibitem[Geiger et~al.(2012)Geiger, Lenz, and Urtasun]{kitti2012}
Andreas Geiger, Philip Lenz, and Raquel Urtasun.
\newblock {Are we ready for Autonomous Driving? {The} {KITTI} {Vision Benchmark
  Suite}}.
\newblock In \emph{CVPR}, 2012.

\bibitem[He et~al.(2025{\natexlab{a}})He, Xu, Guo, Wetzstein, Dai, Li, and
  Yang]{cameractrl2025}
Hao He, Yinghao Xu, Yuwei Guo, Gordon Wetzstein, Bo Dai, Hongsheng Li, and
  Ceyuan Yang.
\newblock {{CameraCtrl}: Enabling Camera Control for Video Diffusion Models}.
\newblock In \emph{ICLR}, 2025{\natexlab{a}}.

\bibitem[He et~al.(2025{\natexlab{b}})He, Li, Yin, Liang, Li, Zhou, Zhang, Liu,
  and Chen]{lotus2025}
Jing He, Haodong Li, Wei Yin, Yixun Liang, Leheng Li, Kaiqiang Zhou, Hongbo
  Zhang, Bingbing Liu, and Ying-Cong Chen.
\newblock {{Lotus}: Diffusion-based Visual Foundation Model for High-quality
  Dense Prediction}.
\newblock In \emph{ICLR}, 2025{\natexlab{b}}.

\bibitem[Hu et~al.(2022)Hu, Shen, Wallis, Allen-Zhu, Li, Wang, Wang, and
  Chen]{lora2022}
Edward~J. Hu, Yelong Shen, Phillip Wallis, Zeyuan Allen-Zhu, Yuanzhi Li, Shean
  Wang, Lu Wang, and Weizhu Chen.
\newblock {{LoRA}: Low-Rank Adaptation of Large Language Models}.
\newblock In \emph{ICLR}, 2022.

\bibitem[Hu et~al.(2025)Hu, Gao, Li, Zhao, Cun, Zhang, Quan, and
  Shan]{depthcrafter2025}
Wenbo Hu, Xiangjun Gao, Xiaoyu Li, Sijie Zhao, Xiaodong Cun, Yong Zhang, Long
  Quan, and Ying Shan.
\newblock {{DepthCrafter}: Generating Consistent Long Depth Sequences for
  Open-world Videos}.
\newblock In \emph{CVPR}, 2025.

\bibitem[Huang et~al.(2026{\natexlab{a}})Huang, Yang, Yang, Ma, Ma, and
  Liao]{gen3r2026}
Jiaxin Huang, Yuanbo Yang, Bangbang Yang, Lin Ma, Yuewen Ma, and Yiyi Liao.
\newblock {{Gen3R}: {3D} Scene Generation Meets Feed-Forward Reconstruction}.
\newblock In \emph{CVPR}, 2026{\natexlab{a}}.

\bibitem[Huang et~al.(2026{\natexlab{b}})Huang, Zhang, He, Gao, Cen, Xia, Zhou,
  Tao, Wan, and Jia]{unityvideo2026}
Jiehui Huang, Yuechen Zhang, Xu He, Yuan Gao, Zhi Cen, Bin Xia, Yan Zhou, Xin
  Tao, Pengfei Wan, and Jiaya Jia.
\newblock {{UnityVideo}: Unified Multi-Modal Multi-Task Learning for Enhancing
  World-Aware Video Generation}.
\newblock In \emph{CVPR}, 2026{\natexlab{b}}.

\bibitem[Huang et~al.(2025)Huang, Zheng, Wang, Liu, Wang, Wu, Jiang, Li, Lau,
  Zuo, and Guo]{voyager2025}
Tianyu Huang, Wangguandong Zheng, Tengfei Wang, Yuhao Liu, Zhenwei Wang, Junta
  Wu, Jie Jiang, Hui Li, Rynson W.~H. Lau, Wangmeng Zuo, and Chunchao Guo.
\newblock {{Voyager}: Long-Range and World-Consistent Video Diffusion for
  Explorable {3D} Scene Generation}.
\newblock \emph{ACM TOG}, 2025.

\bibitem[Huang et~al.(2026{\natexlab{c}})Huang, Li, Lv, and Rehg]{vidfm3d2026}
Zixuan Huang, Xiang Li, Zhaoyang Lv, and James~M. Rehg.
\newblock {How Much {3D} Do Video Foundation Models Encode?}
\newblock In \emph{CVPR}, 2026{\natexlab{c}}.

\bibitem[Jiang et~al.(2025)Jiang, Zheng, Laina, Larlus, and Vedaldi]{geo4d2025}
Zeren Jiang, Chuanxia Zheng, Iro Laina, Diane Larlus, and Andrea Vedaldi.
\newblock {{Geo4D}: Leveraging Video Generators for Geometric {4D} Scene
  Reconstruction}.
\newblock In \emph{ICCV}, 2025.

\bibitem[Karaev et~al.(2023)Karaev, Rocco, Graham, Neverova, Vedaldi, and
  Rupprecht]{dynamicstereo2023}
Nikita Karaev, Ignacio Rocco, Benjamin Graham, Natalia Neverova, Andrea
  Vedaldi, and Christian Rupprecht.
\newblock {{DynamicStereo}: Consistent Dynamic Depth from Stereo Videos}.
\newblock In \emph{CVPR}, 2023.

\bibitem[Ke et~al.(2024)Ke, Obukhov, Huang, Metzger, Caye~Daudt, and
  Schindler]{marigold2024}
Bingxin Ke, Anton Obukhov, Shengyu Huang, Nando Metzger, Rodrigo Caye~Daudt,
  and Konrad Schindler.
\newblock {Repurposing Diffusion-Based Image Generators for Monocular Depth
  Estimation}.
\newblock In \emph{CVPR}, 2024.

\bibitem[Ke et~al.(2025{\natexlab{a}})Ke, Narnhofer, Huang, Ke, Peters,
  Fragkiadaki, Obukhov, and Schindler]{rollingdepth2025}
Bingxin Ke, Dominik Narnhofer, Shengyu Huang, Lei Ke, Torben Peters, Katerina
  Fragkiadaki, Anton Obukhov, and Konrad Schindler.
\newblock {Video Depth without Video Models}.
\newblock In \emph{CVPR}, 2025{\natexlab{a}}.

\bibitem[Ke et~al.(2025{\natexlab{b}})Ke, Qu, Wang, Metzger, Huang, Li,
  Obukhov, and Schindler]{marigoldjournal2025}
Bingxin Ke, Kevin Qu, Tianfu Wang, Nando Metzger, Shengyu Huang, Bo Li, Anton
  Obukhov, and Konrad Schindler.
\newblock {{Marigold}: Affordable Adaptation of Diffusion-Based Image
  Generators for Image Analysis}.
\newblock \emph{TPAMI}, 2025{\natexlab{b}}.

\bibitem[Keetha et~al.(2026)Keetha, M{\"u}ller, Sch{\"o}nberger, Porzi, Zhang,
  Fischer, Knapitsch, Zauss, Weber, Antunes, Luiten, Lopez-Antequera,
  Rota~Bul{\`o}, Richardt, Ramanan, Scherer, and Kontschieder]{mapanything2026}
Nikhil Keetha, Norman M{\"u}ller, Johannes Sch{\"o}nberger, Lorenzo Porzi,
  Yuchen Zhang, Tobias Fischer, Arno Knapitsch, Duncan Zauss, Ethan Weber,
  Nelson Antunes, Jonathon Luiten, Manuel Lopez-Antequera, Samuel
  Rota~Bul{\`o}, Christian Richardt, Deva Ramanan, Sebastian Scherer, and Peter
  Kontschieder.
\newblock {{MapAnything}: Universal Feed-Forward Metric {3D} Reconstruction}.
\newblock In \emph{3DV}, 2026.
\newblock arXiv:2509.13414.

\bibitem[Kingma and Welling(2014)]{vae2013}
Diederik~P. Kingma and Max Welling.
\newblock {Auto-Encoding Variational {Bayes}}.
\newblock In \emph{ICLR}, 2014.

\bibitem[Leroy et~al.(2024)Leroy, Cabon, and Revaud]{mast3r2024}
Vincent Leroy, Yohann Cabon, and J{\'e}r{\^o}me Revaud.
\newblock {Grounding Image Matching in {3D} with {MASt3R}}.
\newblock In \emph{ECCV}, 2024.

\bibitem[Li et~al.(2025)Li, Tucker, Cole, Wang, Jin, Ye, Kanazawa, Holynski,
  and Snavely]{megasam2025}
Zhengqi Li, Richard Tucker, Forrester Cole, Qianqian Wang, Linyi Jin, Vickie
  Ye, Angjoo Kanazawa, Aleksander Holynski, and Noah Snavely.
\newblock {{MegaSaM}: Accurate, Fast and Robust Structure and Motion from
  Casual Dynamic Videos}.
\newblock In \emph{CVPR}, 2025.

\bibitem[Liang et~al.(2025)Liang, Gojcic, Ling, Munkberg, Hasselgren, Lin, Gao,
  Keller, Vijaykumar, Fidler, and Wang]{diffusionrenderer2025}
Ruofan Liang, Zan Gojcic, Huan Ling, Jacob Munkberg, Jon Hasselgren, Zhi-Hao
  Lin, Jun Gao, Alexander Keller, Nandita Vijaykumar, Sanja Fidler, and Zian
  Wang.
\newblock {{DiffusionRenderer}: Neural Inverse and Forward Rendering with Video
  Diffusion Models}.
\newblock In \emph{CVPR}, 2025.

\bibitem[Lin et~al.(2026)Lin, Chen, Liew, Chen, Li, Zhao, Peng, Guo, Zhou, Shi,
  Feng, and Kang]{depthanything3_2026}
Haotong Lin, Sili Chen, Jun~Hao Liew, Donny~Y. Chen, Zhenyu Li, Yang Zhao, Sida
  Peng, Hengkai Guo, Xiaowei Zhou, Guang Shi, Jiashi Feng, and Bingyi Kang.
\newblock {{Depth Anything 3}: Recovering the Visual Space from Any Views}.
\newblock In \emph{ICLR}, 2026.

\bibitem[Ling et~al.(2024)Ling, Sheng, Tu, Zhao, Xin, Wan, Yu, Guo, Yu, Lu, Li,
  Sun, Ashok, Mukherjee, Kang, Kong, Hua, Zhang, Benes, and Bera]{dl3dv2024}
Lu Ling, Yichen Sheng, Zhi Tu, Wentian Zhao, Cheng Xin, Kun Wan, Lantao Yu,
  Qianyu Guo, Zixun Yu, Yawen Lu, Xuanmao Li, Xingpeng Sun, Rohan Ashok,
  Aniruddha Mukherjee, Hao Kang, Xiangrui Kong, Gang Hua, Tianyi Zhang, Bedrich
  Benes, and Aniket Bera.
\newblock {{DL3DV-10K}: A Large-Scale Scene Dataset for Deep Learning-based
  {3D} Vision}.
\newblock In \emph{CVPR}, 2024.

\bibitem[Lipman et~al.(2023)Lipman, Chen, Ben-Hamu, Nickel, and
  Le]{flowmatching2023}
Yaron Lipman, Ricky T.~Q. Chen, Heli Ben-Hamu, Maximilian Nickel, and Matt Le.
\newblock {Flow Matching for Generative Modeling}.
\newblock In \emph{ICLR}, 2023.

\bibitem[Liu et~al.(2026)Liu, Sun, Wang, Wang, Sun, Ye, Zhang, and
  Duan]{reconx2026}
Fangfu Liu, Wenqiang Sun, Hanyang Wang, Yikai Wang, Haowen Sun, Junliang Ye,
  Jun Zhang, and Yueqi Duan.
\newblock {{ReconX}: Reconstruct Any Scene From Sparse Views With Video
  Diffusion Model}.
\newblock \emph{TIP}, 35:\penalty0 2305--2319, 2026.

\bibitem[Liu et~al.(2023)Liu, Gong, and Liu]{rectifiedflow2023}
Xingchao Liu, Chengyue Gong, and Qiang Liu.
\newblock {Flow Straight and Fast: Learning to Generate and Transfer Data with
  Rectified Flow}.
\newblock In \emph{ICLR}, 2023.

\bibitem[Loshchilov and Hutter(2019)]{adamw2019}
Ilya Loshchilov and Frank Hutter.
\newblock {Decoupled Weight Decay Regularization}.
\newblock In \emph{ICLR}, 2019.

\bibitem[Luo et~al.(2020)Luo, Huang, Szeliski, Matzen, and Kopf]{cvd2020}
Xuan Luo, Jia-Bin Huang, Richard Szeliski, Kevin Matzen, and Johannes Kopf.
\newblock {Consistent Video Depth Estimation}.
\newblock \emph{ACM TOG}, 2020.

\bibitem[Mai et~al.(2025)Mai, Zhu, Liu, Li, Zheng, Schmidhuber, and
  Ghanem]{sora3r2025}
Jinjie Mai, Wenxuan Zhu, Haozhe Liu, Bing Li, Cheng Zheng, J{\"u}rgen
  Schmidhuber, and Bernard Ghanem.
\newblock {Can Video Diffusion Model Reconstruct {4D} Geometry?}
\newblock \emph{arXiv preprint arXiv:2503.21082}, 2025.

\bibitem[Man et~al.(2024)Man, Zheng, Bao, Hebert, Gui, and Wang]{lexicon3d2024}
Yunze Man, Shuhong Zheng, Zhipeng Bao, Martial Hebert, Liang-Yan Gui, and
  Yu-Xiong Wang.
\newblock {{Lexicon3D}: Probing Visual Foundation Models for Complex {3D} Scene
  Understanding}.
\newblock In \emph{NeurIPS}, 2024.

\bibitem[Martin~Garcia et~al.(2025)Martin~Garcia, Knaebel, Schmidt, de~Geus,
  Hermans, and Leibe]{e2eft2025}
Gonzalo Martin~Garcia, Karim Knaebel, Christian Schmidt, Daan de Geus,
  Alexander Hermans, and Bastian Leibe.
\newblock {Fine-Tuning Image-Conditional Diffusion Models is Easier than You
  Think}.
\newblock In \emph{WACV}, 2025.

\bibitem[Mehl et~al.(2023)Mehl, Schmalfuss, Jahedi, Nalivayko, and
  Bruhn]{spring2023}
Lukas Mehl, Jenny Schmalfuss, Azin Jahedi, Yaroslava Nalivayko, and Andr{\'e}s
  Bruhn.
\newblock {Spring: A High-Resolution High-Detail Dataset and Benchmark for
  Scene Flow, Optical Flow and Stereo}.
\newblock In \emph{CVPR}, 2023.

\bibitem[Oquab et~al.(2024)Oquab, Darcet, Moutakanni, Vo, Szafraniec, Khalidov,
  Fernandez, Haziza, Massa, El-Nouby, Assran, Ballas, Galuba, Howes, Huang, Li,
  Misra, Rabbat, Sharma, Synnaeve, Xu, J{\'e}gou, Mairal, Labatut, Joulin, and
  Bojanowski]{dinov2_2024}
Maxime Oquab, Timoth{\'e}e Darcet, Th{\'e}o Moutakanni, Huy Vo, Marc
  Szafraniec, Vasil Khalidov, Pierre Fernandez, Daniel Haziza, Francisco Massa,
  Alaaeldin El-Nouby, Mahmoud Assran, Nicolas Ballas, Wojciech Galuba, Russell
  Howes, Po-Yao Huang, Shang-Wen Li, Ishan Misra, Michael Rabbat, Vasu Sharma,
  Gabriel Synnaeve, Hu Xu, Herv{\'e} J{\'e}gou, Julien Mairal, Patrick Labatut,
  Armand Joulin, and Piotr Bojanowski.
\newblock {{DINOv2}: Learning Robust Visual Features without Supervision}.
\newblock \emph{TMLR}, 2024.

\bibitem[Palazzolo et~al.(2019)Palazzolo, Behley, Lottes, Gigu{\`e}re, and
  Stachniss]{bonn2019}
Emanuele Palazzolo, Jens Behley, Philipp Lottes, Philippe Gigu{\`e}re, and
  Cyrill Stachniss.
\newblock {{ReFusion}: {3D} Reconstruction in Dynamic Environments for {RGB-D}
  Cameras Exploiting Residuals}.
\newblock In \emph{IROS}, 2019.

\bibitem[Peebles and Xie(2023)]{dit2023}
William Peebles and Saining Xie.
\newblock {Scalable Diffusion Models with Transformers}.
\newblock In \emph{ICCV}, 2023.

\bibitem[Ren et~al.(2025)Ren, Shen, Huang, Ling, Lu, Nimier-David, M{\"u}ller,
  Keller, Fidler, and Gao]{gen3c2025}
Xuanchi Ren, Tianchang Shen, Jiahui Huang, Huan Ling, Yifan Lu, Merlin
  Nimier-David, Thomas M{\"u}ller, Alexander Keller, Sanja Fidler, and Jun Gao.
\newblock {{GEN3C}: {3D}-Informed World-Consistent Video Generation with
  Precise Camera Control}.
\newblock In \emph{CVPR}, 2025.

\bibitem[Rombach et~al.(2022)Rombach, Blattmann, Lorenz, Esser, and
  Ommer]{rombach2022ldm}
Robin Rombach, Andreas Blattmann, Dominik Lorenz, Patrick Esser, and Bj{\"o}rn
  Ommer.
\newblock {High-Resolution Image Synthesis with Latent Diffusion Models}.
\newblock In \emph{CVPR}, 2022.

\bibitem[Sch{\"o}ps et~al.(2017)Sch{\"o}ps, Sch{\"o}nberger, Galliani, Sattler,
  Schindler, Pollefeys, and Geiger]{eth3d2017}
Thomas Sch{\"o}ps, Johannes~L. Sch{\"o}nberger, Silvano Galliani, Torsten
  Sattler, Konrad Schindler, Marc Pollefeys, and Andreas Geiger.
\newblock {A Multi-View Stereo Benchmark with High-Resolution Images and
  Multi-Camera Videos}.
\newblock In \emph{CVPR}, 2017.

\bibitem[Sch{\"o}ps et~al.(2019)Sch{\"o}ps, Sattler, and
  Pollefeys]{badslam2019}
Thomas Sch{\"o}ps, Torsten Sattler, and Marc Pollefeys.
\newblock {{BAD SLAM}: {Bundle Adjusted Direct} {RGB-D} {SLAM}}.
\newblock In \emph{CVPR}, 2019.

\bibitem[Shao et~al.(2025)Shao, Yang, Zhou, Zhang, Shen, Guizilini, Wang,
  Poggi, and Liao]{chronodepth2025}
Jiahao Shao, Yuanbo Yang, Hongyu Zhou, Youmin Zhang, Yujun Shen, Vitor
  Guizilini, Yue Wang, Matteo Poggi, and Yiyi Liao.
\newblock {Learning Temporally Consistent Video Depth from Video Diffusion
  Priors}.
\newblock In \emph{CVPR}, 2025.

\bibitem[Shen et~al.(2026{\natexlab{a}})Shen, Zhao, Zhao, and
  Yin]{probevlmvgm2026}
Haozhan Shen, Tiancheng Zhao, Kangjia Zhao, and Jianwei Yin.
\newblock {Which Pretraining Paradigm Better Serves Spatial Intelligence? {An}
  Empirical Comparison of Vision-Language and Video Generation Models}.
\newblock \emph{arXiv preprint arXiv:2605.28132}, 2026{\natexlab{a}}.

\bibitem[Shen et~al.(2026{\natexlab{b}})Shen, Bahmani, He, Srinivasan, Cao,
  Ren, Li, Wang, Sharp, Gojcic, Fidler, Huang, Ling, Gao, and Ren]{lyra2_2026}
Tianchang Shen, Sherwin Bahmani, Kai He, Sangeetha~Grama Srinivasan, Tianshi
  Cao, Jiawei Ren, Ruilong Li, Zian Wang, Nicholas Sharp, Zan Gojcic, Sanja
  Fidler, Jiahui Huang, Huan Ling, Jun Gao, and Xuanchi Ren.
\newblock {{Lyra 2.0}: Explorable Generative {3D} Worlds}.
\newblock \emph{arXiv preprint arXiv:2604.13036}, 2026{\natexlab{b}}.

\bibitem[Shotton et~al.(2013)Shotton, Glocker, Zach, Izadi, Criminisi, and
  Fitzgibbon]{sevenscenes2013}
Jamie Shotton, Ben Glocker, Christopher Zach, Shahram Izadi, Antonio Criminisi,
  and Andrew Fitzgibbon.
\newblock {Scene Coordinate Regression Forests for Camera Relocalization in
  {RGB-D} Images}.
\newblock In \emph{CVPR}, 2013.

\bibitem[Sturm et~al.(2012)Sturm, Engelhard, Endres, Burgard, and
  Cremers]{tumrgbd2012}
J{\"u}rgen Sturm, Nikolas Engelhard, Felix Endres, Wolfram Burgard, and Daniel
  Cremers.
\newblock {A Benchmark for the Evaluation of {RGB-D} {SLAM} Systems}.
\newblock In \emph{IROS}, 2012.

\bibitem[Sun et~al.(2020)Sun, Kretzschmar, Dotiwalla, Chouard, Patnaik, Tsui,
  Guo, Zhou, Chai, Caine, Vasudevan, Han, Ngiam, Zhao, Timofeev, Ettinger,
  Krivokon, Gao, Joshi, Zhang, Shlens, Chen, and Anguelov]{waymo2020}
Pei Sun, Henrik Kretzschmar, Xerxes Dotiwalla, Aurelien Chouard, Vijaysai
  Patnaik, Paul Tsui, James Guo, Yin Zhou, Yuning Chai, Benjamin Caine, Vijay
  Vasudevan, Wei Han, Jiquan Ngiam, Hang Zhao, Aleksei Timofeev, Scott
  Ettinger, Maxim Krivokon, Amy Gao, Aditya Joshi, Yu Zhang, Jonathon Shlens,
  Zhifeng Chen, and Dragomir Anguelov.
\newblock {Scalability in Perception for Autonomous Driving: {Waymo} {Open
  Dataset}}.
\newblock In \emph{CVPR}, 2020.

\bibitem[{Team Wan} et~al.(2025){Team Wan}, Wang, Ai, Wen, Mao, Xie, Chen, Yu,
  Zhao, Yang, Zeng, Wang, Zhang, Zhou, Wang, Chen, Zhu, Zhao, Yan, Huang, Feng,
  Zhang, Li, Wu, Chu, Feng, Zhang, Sun, Fang, Wang, Gui, Weng, Shen, Lin, Wang,
  Wang, Zhou, Wang, Shen, Yu, Shi, Huang, Xu, Kou, Lv, Li, Liu, Wang, Zhang,
  Huang, Li, Wu, Liu, Pan, Zheng, Hong, Shi, Feng, Jiang, Han, Wu, and
  Liu]{wan2025}
{Team Wan}, Ang Wang, Baole Ai, Bin Wen, Chaojie Mao, Chen-Wei Xie, Di Chen,
  Feiwu Yu, Haiming Zhao, Jianxiao Yang, Jianyuan Zeng, Jiayu Wang, Jingfeng
  Zhang, Jingren Zhou, Jinkai Wang, Jixuan Chen, Kai Zhu, Kang Zhao, Keyu Yan,
  Lianghua Huang, Mengyang Feng, Ningyi Zhang, Pandeng Li, Pingyu Wu, Ruihang
  Chu, Ruili Feng, Shiwei Zhang, Siyang Sun, Tao Fang, Tianxing Wang, Tianyi
  Gui, Tingyu Weng, Tong Shen, Wei Lin, Wei Wang, Wei Wang, Wenmeng Zhou, Wente
  Wang, Wenting Shen, Wenyuan Yu, Xianzhong Shi, Xiaoming Huang, Xin Xu, Yan
  Kou, Yangyu Lv, Yifei Li, Yijing Liu, Yiming Wang, Yingya Zhang, Yitong
  Huang, Yong Li, You Wu, Yu Liu, Yulin Pan, Yun Zheng, Yuntao Hong, Yupeng
  Shi, Yutong Feng, Zeyinzi Jiang, Zhen Han, Zhi-Fan Wu, and Ziyu Liu.
\newblock {{Wan}: Open and Advanced Large-Scale Video Generative Models}.
\newblock \emph{arXiv preprint arXiv:2503.20314}, 2025.

\bibitem[Uhrig et~al.(2017)Uhrig, Schneider, Schneider, Franke, Brox, and
  Geiger]{kittidepth2017}
Jonas Uhrig, Nick Schneider, Lukas Schneider, Uwe Franke, Thomas Brox, and
  Andreas Geiger.
\newblock {Sparsity Invariant {CNNs}}.
\newblock In \emph{3DV}, 2017.

\bibitem[Voleti et~al.(2024)Voleti, Yao, Boss, Letts, Pankratz, Tochilkin,
  Laforte, Rombach, and Jampani]{sv3d2024}
Vikram Voleti, Chun-Han Yao, Mark Boss, Adam Letts, David Pankratz, Dmitrii
  Tochilkin, Christian Laforte, Robin Rombach, and Varun Jampani.
\newblock {{SV3D}: Novel Multi-view Synthesis and {3D} Generation from a Single
  Image using Latent Video Diffusion}.
\newblock In \emph{ECCV}, 2024.

\bibitem[Wang et~al.(2025{\natexlab{a}})Wang, Chen, Karaev, Vedaldi, Rupprecht,
  and Novotny]{vggt2025}
Jianyuan Wang, Minghao Chen, Nikita Karaev, Andrea Vedaldi, Christian
  Rupprecht, and David Novotny.
\newblock {{VGGT}: {Visual Geometry Grounded Transformer}}.
\newblock In \emph{CVPR}, 2025{\natexlab{a}}.

\bibitem[Wang et~al.(2026{\natexlab{a}})Wang, Chen, Zhang, Karaev,
  Sch{\"o}nberger, Labatut, Bojanowski, Novotny, Vedaldi, and
  Rupprecht]{vggtomega2026}
Jianyuan Wang, Minghao Chen, Shangzhan Zhang, Nikita Karaev, Johannes
  Sch{\"o}nberger, Patrick Labatut, Piotr Bojanowski, David Novotny, Andrea
  Vedaldi, and Christian Rupprecht.
\newblock {{VGGT}-{$\Omega$}}.
\newblock In \emph{CVPR}, 2026{\natexlab{a}}.
\newblock arXiv:2605.15195.

\bibitem[Wang et~al.(2026{\natexlab{b}})Wang, Zhang, Kabra, Uijlings,
  Waslander, Zisserman, Carreira, He, Andriluka, Bazavan, Zanfir, and
  Sminchisescu]{genception2026}
Letian Wang, Chuhan Zhang, Rishabh Kabra, Jasper Uijlings, Steven Waslander,
  Andrew Zisserman, Joao Carreira, Kaiming He, Misha Andriluka, Eduard~Gabriel
  Bazavan, Andrei Zanfir, and Cristian Sminchisescu.
\newblock {Video Generation Models Are General-Purpose Vision Learners}.
\newblock In \emph{ECCV}, 2026{\natexlab{b}}.

\bibitem[Wang et~al.(2025{\natexlab{b}})Wang, Zhang, Holynski, Efros, and
  Kanazawa]{cut3r2025}
Qianqian Wang, Yifei Zhang, Aleksander Holynski, Alexei~A. Efros, and Angjoo
  Kanazawa.
\newblock {Continuous {3D} Perception Model with Persistent State}.
\newblock In \emph{CVPR}, 2025{\natexlab{b}}.

\bibitem[Wang et~al.(2025{\natexlab{c}})Wang, Xu, Dai, Xiang, Deng, Tong, and
  Yang]{moge2025}
Ruicheng Wang, Sicheng Xu, Cassie Dai, Jianfeng Xiang, Yu Deng, Xin Tong, and
  Jiaolong Yang.
\newblock {{MoGe}: Unlocking Accurate Monocular Geometry Estimation for
  Open-Domain Images with Optimal Training Supervision}.
\newblock In \emph{CVPR}, 2025{\natexlab{c}}.

\bibitem[Wang et~al.(2025{\natexlab{d}})Wang, Xu, Dong, Deng, Xiang, Lv, Sun,
  Tong, and Yang]{moge2_2025}
Ruicheng Wang, Sicheng Xu, Yue Dong, Yu Deng, Jianfeng Xiang, Zelong Lv,
  Guangzhong Sun, Xin Tong, and Jiaolong Yang.
\newblock {{MoGe-2}: Accurate Monocular Geometry with Metric Scale and Sharp
  Details}.
\newblock In \emph{NeurIPS}, 2025{\natexlab{d}}.

\bibitem[Wang et~al.(2024)Wang, Leroy, Cabon, Chidlovskii, and
  Revaud]{dust3r2024}
Shuzhe Wang, Vincent Leroy, Yohann Cabon, Boris Chidlovskii, and Jerome Revaud.
\newblock {{DUSt3R}: Geometric {3D} Vision Made Easy}.
\newblock In \emph{CVPR}, 2024.

\bibitem[Wang et~al.(2020)Wang, Zhu, Wang, Hu, Qiu, Wang, Hu, Kapoor, and
  Scherer]{tartanair2020}
Wenshan Wang, Delong Zhu, Xiangwei Wang, Yaoyu Hu, Yuheng Qiu, Chen Wang, Yafei
  Hu, Ashish Kapoor, and Sebastian Scherer.
\newblock {{TartanAir}: A Dataset to Push the Limits of Visual {SLAM}}.
\newblock In \emph{IROS}, 2020.

\bibitem[Wang et~al.(2023)Wang, Hu, Qiu, Shen, and Shaoul]{tartanairv2}
Wenshan Wang, Yaoyu Hu, Yuheng Qiu, Shihao Shen, and Yorai Shaoul.
\newblock {{TartanAir} Dataset Documentation}.
\newblock \url{https://tartanair.org/}, 2023.
\newblock TartanAir V2. Accessed 2026-10-25.

\bibitem[Wang et~al.(2026{\natexlab{c}})Wang, Zhou, Zhu, Chang, Zhou, Li, Chen,
  Pang, Shen, and He]{pi3_2026}
Yifan Wang, Jianjun Zhou, Haoyi Zhu, Wenzheng Chang, Yang Zhou, Zizun Li, Junyi
  Chen, Jiangmiao Pang, Chunhua Shen, and Tong He.
\newblock {$\pi^3$: Permutation-Equivariant Visual Geometry Learning}.
\newblock In \emph{ICLR}, 2026{\natexlab{c}}.

\bibitem[Wu et~al.(2026)Wu, Wu, He, Guo, Ye, Duan, and
  Bian]{geometryforcing2026}
Haoyu Wu, Diankun Wu, Tianyu He, Junliang Guo, Yang Ye, Yueqi Duan, and Jiang
  Bian.
\newblock {{Geometry Forcing}: Marrying Video Diffusion and {3D} Representation
  for Consistent World Modeling}.
\newblock In \emph{ICLR}, 2026.

\bibitem[Xi et~al.(2026)Xi, Wang, Liang, Qiu, Huo, Wang, Zhang, and
  Li]{omnivdiff2026}
Dianbing Xi, Jiepeng Wang, Yuanzhi Liang, Xi Qiu, Yuchi Huo, Rui Wang, Chi
  Zhang, and Xuelong Li.
\newblock {{OmniVDiff}: Omni Controllable Video Diffusion for Generation and
  Understanding}.
\newblock In \emph{AAAI}, 2026.

\bibitem[Xia et~al.(2026)Xia, Jiang, and Zheng]{gen4p2026}
Yitong Xia, Zeren Jiang, and Chuanxia Zheng.
\newblock {{Gen4P}: Video Generators are {4D} Perceivers}.
\newblock \emph{arXiv preprint arXiv:2610.22371}, 2026.

\bibitem[Xie et~al.(2025)Xie, Yao, Voleti, Jiang, and Jampani]{sv4d2025}
Yiming Xie, Chun-Han Yao, Vikram Voleti, Huaizu Jiang, and Varun Jampani.
\newblock {{SV4D}: Dynamic {3D} Content Generation with Multi-Frame and
  Multi-View Consistency}.
\newblock In \emph{ICLR}, 2025.

\bibitem[Xu et~al.(2025{\natexlab{a}})Xu, Ge, Liu, Fan, Xie, Zhao, Chen, and
  Shen]{genpercept2025}
Guangkai Xu, Yongtao Ge, Mingyu Liu, Chengxiang Fan, Kangyang Xie, Zhiyue Zhao,
  Hao Chen, and Chunhua Shen.
\newblock {What Matters When Repurposing Diffusion Models for General Dense
  Perception Tasks?}
\newblock In \emph{ICLR}, 2025{\natexlab{a}}.

\bibitem[Xu et~al.(2026)Xu, Wei, Wei, Geng, Li, Shen, Sun, Han, Ma, Li, Ye,
  Zheng, Wang, Zhang, and Zhao]{dkt2025}
Shaocong Xu, Songlin Wei, Qizhe Wei, Zheng Geng, Hong Li, Licheng Shen, Qianpu
  Sun, Shu Han, Bin Ma, Bohan Li, Chongjie Ye, Yuhang Zheng, Nan Wang, Saining
  Zhang, and Hao Zhao.
\newblock {{Diffusion Knows Transparency}: Repurposing Video Diffusion for
  Transparent Object Depth and Normal Estimation}.
\newblock In \emph{ICRA}, 2026.

\bibitem[Xu et~al.(2025{\natexlab{b}})Xu, Gao, Hu, Li, Zhang, and
  Shan]{geometrycrafter2025}
Tian-Xing Xu, Xiangjun Gao, Wenbo Hu, Xiaoyu Li, Song-Hai Zhang, and Ying Shan.
\newblock {{GeometryCrafter}: Consistent Geometry Estimation for Open-world
  Videos with Diffusion Priors}.
\newblock In \emph{ICCV}, 2025{\natexlab{b}}.

\bibitem[Yang et~al.(2025{\natexlab{a}})Yang, Huang, Yin, Shen, Liu, He, Lin,
  Ouyang, and He]{depthanyvideo2025}
Honghui Yang, Di Huang, Wei Yin, Chunhua Shen, Haifeng Liu, Xiaofei He, Binbin
  Lin, Wanli Ouyang, and Tong He.
\newblock {{Depth Any Video} with Scalable Synthetic Data}.
\newblock In \emph{ICLR}, 2025{\natexlab{a}}.

\bibitem[Yang et~al.(2026)Yang, Song, Zhang, Zhu, and Deng]{geonext2026}
Haosen Yang, Jifei Song, Zhensong Zhang, Xiatian Zhu, and Jiankang Deng.
\newblock {Video Generative Models as Geometry Learner}.
\newblock In \emph{ECCV}, 2026.

\bibitem[Yang et~al.(2024)Yang, Kang, Huang, Zhao, Xu, Feng, and
  Zhao]{depthanythingv2_2024}
Lihe Yang, Bingyi Kang, Zilong Huang, Zhen Zhao, Xiaogang Xu, Jiashi Feng, and
  Hengshuang Zhao.
\newblock {Depth Anything V2}.
\newblock In \emph{NeurIPS}, 2024.

\bibitem[Yang et~al.(2025{\natexlab{b}})Yang, Teng, Zheng, Ding, Huang, Xu,
  Yang, Hong, Zhang, Feng, Yin, Zhang, Wang, Cheng, Xu, Gu, Dong, and
  Tang]{cogvideox2025}
Zhuoyi Yang, Jiayan Teng, Wendi Zheng, Ming Ding, Shiyu Huang, Jiazheng Xu,
  Yuanming Yang, Wenyi Hong, Xiaohan Zhang, Guanyu Feng, Da Yin, Yuxuan Zhang,
  Weihan Wang, Yean Cheng, Bin Xu, Xiaotao Gu, Yuxiao Dong, and Jie Tang.
\newblock {{CogVideoX}: Text-to-Video Diffusion Models with An Expert
  Transformer}.
\newblock In \emph{ICLR}, 2025{\natexlab{b}}.

\bibitem[Yeshwanth et~al.(2023)Yeshwanth, Liu, Nie{\ss}ner, and
  Dai]{scannetpp2023}
Chandan Yeshwanth, Yueh-Cheng Liu, Matthias Nie{\ss}ner, and Angela Dai.
\newblock {{ScanNet++}: A High-Fidelity Dataset of {3D} Indoor Scenes}.
\newblock In \emph{ICCV}, 2023.

\bibitem[Yin et~al.(2021)Yin, Zhang, Wang, Niklaus, Mai, Chen, and
  Shen]{leres2021}
Wei Yin, Jianming Zhang, Oliver Wang, Simon Niklaus, Long Mai, Simon Chen, and
  Chunhua Shen.
\newblock {Learning to Recover {3D} Scene Shape from a Single Image}.
\newblock In \emph{CVPR}, 2021.

\bibitem[Yu et~al.(2025)Yu, Xing, Yuan, Hu, Li, Huang, Gao, Wong, Shan, and
  Tian]{viewcrafter2025}
Wangbo Yu, Jinbo Xing, Li Yuan, Wenbo Hu, Xiaoyu Li, Zhipeng Huang, Xiangjun
  Gao, Tien-Tsin Wong, Ying Shan, and Yonghong Tian.
\newblock {{ViewCrafter}: Taming Video Diffusion Models for High-fidelity Novel
  View Synthesis}.
\newblock \emph{TPAMI}, 2025.

\bibitem[Zhang et~al.(2026{\natexlab{a}})Zhang, Chen, Liao, He, Zhang, Li,
  Liang, Chen, Ren, Zheng, Yang, Zhou, Li, Sebe, and Chen]{dvd2026}
Hongfei Zhang, Harold~Haodong Chen, Chenfei Liao, Jing He, Zixin Zhang, Haodong
  Li, Yihao Liang, Kanghao Chen, Bin Ren, Xu Zheng, Shuai Yang, Kun Zhou,
  Yinchuan Li, Nicu Sebe, and Ying-Cong Chen.
\newblock {{DVD}: Deterministic Video Depth Estimation with Generative Priors}.
\newblock \emph{arXiv preprint arXiv:2603.12250}, 2026{\natexlab{a}}.

\bibitem[Zhang et~al.(2025)Zhang, Herrmann, Hur, Jampani, Darrell, Cole, Sun,
  and Yang]{monst3r2025}
Junyi Zhang, Charles Herrmann, Junhwa Hur, Varun Jampani, Trevor Darrell,
  Forrester Cole, Deqing Sun, and Ming-Hsuan Yang.
\newblock {{MonST3R}: A Simple Approach for Estimating Geometry in the Presence
  of Motion}.
\newblock In \emph{ICLR}, 2025.

\bibitem[Zhang et~al.(2026{\natexlab{b}})Zhang, Ouyang, Chen, Mei, Chen, Ren,
  Sun, Zhang, Liu, and Wang]{latentbridge2026}
Shuai Zhang, Zhanpeng Ouyang, Zhengqing Chen, Ruohong Mei, Shaoyu Chen,
  Weiqiang Ren, Hui Sun, Qian Zhang, Wenyu Liu, and Xinggang Wang.
\newblock {{Latent Bridge}: From Video Latents to {3D} Geometry via Feature
  Alignment}.
\newblock \emph{arXiv preprint arXiv:2610.28521}, 2026{\natexlab{b}}.

\bibitem[Zheng et~al.(2023)Zheng, Harley, Shen, Wetzstein, and
  Guibas]{pointodyssey2023}
Yang Zheng, Adam~W. Harley, Bokui Shen, Gordon Wetzstein, and Leonidas~J.
  Guibas.
\newblock {{PointOdyssey}: A Large-Scale Synthetic Dataset for Long-Term Point
  Tracking}.
\newblock In \emph{ICCV}, 2023.

\bibitem[Zhou et~al.(2018)Zhou, Tucker, Flynn, Fyffe, and
  Snavely]{realestate10k2018}
Tinghui Zhou, Richard Tucker, John Flynn, Graham Fyffe, and Noah Snavely.
\newblock {Stereo Magnification: Learning View Synthesis using Multiplane
  Images}.
\newblock \emph{ACM TOG}, 2018.

\bibitem[Zhu et~al.(2025)Zhu, Wang, Zhou, Chang, Zhou, Li, Chen, Shen, Pang,
  and He]{aether2025}
Haoyi Zhu, Yifan Wang, Jianjun Zhou, Wenzheng Chang, Yang Zhou, Zizun Li, Junyi
  Chen, Chunhua Shen, Jiangmiao Pang, and Tong He.
\newblock {{Aether}: Geometric-Aware Unified World Modeling}.
\newblock In \emph{ICCV}, 2025.

\bibitem[Zhu et~al.(2026)Zhu, Lu, Hu, Han, Cai, Shan, and
  Zheng]{motioncrafter2026}
Ruijie Zhu, Jiahao Lu, Wenbo Hu, Xiaoguang Han, Jianfei Cai, Ying Shan, and
  Chuanxia Zheng.
\newblock {{MotionCrafter}: Dense Geometry and Motion Reconstruction with a
  {4D} {VAE}}.
\newblock In \emph{CVPR}, 2026.

\end{thebibliography}
}
\end{document}